%% file: main.tex
\documentclass{article}

\input{UTILS/PreSetting}
\input{UTILS/math_commands}


\newcommand{\xhdr}[1]{\noindent{{\bf #1.}}}

\newcommand{\hide}[1]{}
\usepackage{tabularx}
\usepackage{ragged2e}
\newcolumntype{L}[1]{>{\RaggedRight\arraybackslash}p{#1}}

\title{\LARGE\textbf{Empowering Biomedical Discovery with AI Agents}}

\author[1]{Shanghua Gao}
\author[1,2,8,+]{Ada Fang}
\author[1,3,+]{Yepeng Huang}
\author[1,4,+]{Valentina Giunchiglia}
\author[1,5,+]{Ayush Noori}
\author[1]{Jonathan Richard Schwarz}
\author[1,6]{Yasha Ektefaie}
\author[7]{Jovana Kondic}
\author[1,8,9,10,\#]{Marinka Zitnik}
\affil[1]{Department of Biomedical Informatics, Harvard Medical School, Boston, MA, USA}
\affil[2]{Department of Chemistry and Chemical Biology, Harvard University, Cambridge, MA, USA}
\affil[3]{Program in Biological and Biomedical Sciences, Harvard Medical School, Boston, MA, USA}
\affil[4]{Department of Brain Sciences, Imperial College London, London, UK}
\affil[5]{Harvard College, Cambridge, MA, USA}
\affil[6]{Program in Biomedical Informatics, Harvard Medical School, Boston, MA, USA}
\affil[7]{Department of Electrical Engineering and Computer Science, MIT, Cambridge, MA, USA}
\affil[8]{Kempner Institute for the Study of Natural and Artificial Intelligence, Harvard University, MA, USA}
\affil[9]{Broad Institute of MIT and Harvard, Cambridge, MA, USA}
\affil[10]{Harvard Data Science Initiative, Cambridge, MA, USA}

\affil[+]{\em{Co-second authors}}
\affil[$\#$]{\em{Correspondence: \href{mailto:marinka@hms.harvard.edu}{marinka@hms.harvard.edu}}}

\date{}
\begin{document}

\maketitle
\renewcommand{\abstractname}{Perspective Summary}
\captionsetup[table]{name=Table}
\begin{abstract}
	\input{00abstract}
\end{abstract}

\input{01intro}
\input{02why}
\input{03how}

\input{04levels}
\input{05components}

\input{06challenges}
\input{07outlook}

\clearpage

\paragraph{Declaration of interests}

The authors declare no competing interests.

\paragraph{Acknowledgments}

We gratefully acknowledge the support of NIH R01-HD108794, NSF CAREER 2339524, US DoD FA8702-15-D-0001, awards from Harvard Data Science Initiative, Amazon Faculty Research, Google Research Scholar Program, AstraZeneca Research, Roche Alliance with Distinguished Scientists, Sanofi iDEA-iTECH Award, Pfizer Research, Chan Zuckerberg Initiative, John and Virginia Kaneb Fellowship award at Harvard Medical School, Aligning Science Across Parkinson's (ASAP) Initiative, Biswas Computational Biology Initiative in partnership with the Milken Institute, and Kempner Institute for the Study of Natural and Artificial Intelligence at Harvard University. A.F. is supported by the Kempner Institute Graduate Fellowship. A.N. is supported by the Herchel Smith-Harvard Undergraduate Science Fellowship, the Yun Family Research Fellows Fund for Revolutionary Thinking, and the Summer Institute in Biomedical Informatics at Harvard Medical School. V.G. is supported by the Medical Research Council, MR/W00710X/1. Y.E. is supported by grant T32 HG002295 from the National Human Genome Research Institute and the NSDEG fellowship. The authors would like to thank Owen Queen, Alejandro Velez-Arce, and Ruth Johnson for their constructive comments on the draft manuscript. Any opinions, findings, conclusions or recommendations expressed in this material are those of the authors and do not necessarily reflect the views of the funders.

\paragraph{Author contributions}

All authors contributed to the design and writing of the manuscript, helped shape the research, provided critical feedback, and commented on the manuscript and its revisions. M.Z. conceived the study and was in charge of overall direction and planning.

\clearpage

\clearpage

\input{08figure}

\clearpage

\input{09box}

\clearpage

\citestyle{nature}
\bibliographystyle{unsrtnat} 
\bibliography{ref}

\end{document}

%% file: UTILS/PreSetting.tex
\usepackage[12pt]{extsizes}
\usepackage[utf8]{inputenc}
\usepackage{amsmath}
\usepackage{amssymb}
\usepackage{amsthm}
\usepackage{amsfonts}
\usepackage[numbers,sort&compress]{natbib}
\usepackage{graphicx}
\usepackage{lipsum}
\usepackage{enumitem}
\usepackage[font=footnotesize,labelfont=bf]{caption}
\usepackage{mdframed}
\usepackage{authblk}
\usepackage{longtable}
\usepackage{subfigure}
\usepackage[subfigure]{tocloft}
\usepackage[dvipsnames,table]{xcolor}
\usepackage[hidelinks]{hyperref}
\usepackage{mathtools}
\usepackage{relsize}
\usepackage{multirow}
\usepackage{booktabs}
\usepackage{tabularx}
\usepackage{bbding}
\usepackage{listings}
\usepackage{xspace}
\usepackage[ruled,vlined]{algorithm2e}
\usepackage{multirow}
\usepackage[p,osf]{cochineal}
\usepackage[scale=.95,type1]{cabin}
\usepackage[cochineal,bigdelims,cmintegrals,vvarbb]{newtxmath}
\usepackage[zerostyle=c,scaled=.94]{newtxtt}
\usepackage[cal=boondoxo]{mathalfa}
\usepackage{microtype}
\usepackage{multirow}
\usepackage{adjustbox}
\usepackage{etoolbox}
\usepackage{lmodern}
\usepackage{csquotes}

\usepackage{caption}
\usepackage{geometry}
\definecolor{myurlcolor}{HTML}{123463}
\definecolor{dc_color}{RGB}{230, 245, 244}
\definecolor{ds_color}{RGB}{195, 230, 227}
\definecolor{ms_color}{RGB}{150, 214, 209}
\hypersetup{
    colorlinks=true,
    linkcolor=black,
    filecolor=black,      
    urlcolor=black,
    citecolor=black
}

\newcommand{\eg}{\emph{e.g.}\xspace}

\apptocmd{\thebibliography}{\raggedright}{}{}

\makeatletter
\patchcmd{\@maketitle}{\LARGE \@title}{\fontsize{30}{19.2}\selectfont\@title}{}{}
\makeatother
\usepackage{cleveref}
\Crefname{section}{Sec.}{Secs.}
\Crefname{equation}{Eq.}{Eqs.}
\Crefname{figure}{Fig.}{Figs.}
\Crefname{tabular}{Tab.}{Tabs.}

%% file: UTILS/math_commands.tex
\usepackage{amsmath,amsfonts,bm}

\def\eqref#1{equation~\ref{#1}}

\DeclareMathAlphabet{\mathsfit}{\encodingdefault}{\sfdefault}{m}{sl}
\SetMathAlphabet{\mathsfit}{bold}{\encodingdefault}{\sfdefault}{bx}{n}

%% file: 00abstract.tex
\noindent We envision ``AI scientists'' as systems capable of skeptical learning and reasoning that empower biomedical research through collaborative agents that integrate AI models and biomedical tools with experimental platforms.
Rather than taking humans out of the discovery process, biomedical AI agents combine human creativity and expertise with AI's ability to analyze large datasets, navigate hypothesis spaces, and execute repetitive tasks.
AI agents are poised to be proficient in various tasks, planning discovery workflows and performing self-assessment to identify and mitigate gaps in their knowledge. These agents use large language models and generative models to feature structured memory for continual learning and use machine learning tools to incorporate scientific knowledge, biological principles, and theories.
AI agents can impact areas ranging from virtual cell simulation, programmable control of phenotypes, and the design of cellular circuits to developing new therapies.

%% file: 01intro.tex
\section*{Introduction}
A long-standing ambition for artificial intelligence (AI) is the development of AI systems that can eventually make major scientific discoveries, learn on their own, and acquire knowledge autonomously.
While this concept of an ``AI scientist'' is aspirational, advances in agent-based AI pave the way to the development of {\em AI agents} as conversable systems capable of reflective learning and reasoning that coordinate large language models (LLMs), machine learning (ML) tools, experimental platforms, or even combinations of them~\cite{boiko2023autonomous,bran2023chemcrow,xi2023rise,guo2024large} (Figure~\ref{fig:fig1}). The complexity of biological problems requires an approach where a complex problem is decomposed into simpler tasks. AI agents can break down a problem into manageable subtasks, which can then be addressed by agents with specialized functions for targeted problem-solving and integration of scientific knowledge~\cite{boiko2023autonomous,wang2023scientific}.
In the near future, AI agents can accelerate discovery workflows by making them faster and more resource-efficient. AI agents improve the efficiency of routine tasks, automate repetitive processes, and analyze large datasets to navigate hypothesis spaces at a scale and precision that surpasses current human-driven efforts. This automation allows for continuous, high-throughput research that would be impossible for human researchers to perform alone at the same scale or speed. Looking further ahead, AI agents can enable insights that might not have been possible using ML alone by making predictions across temporal and spatial scales prior to experimental measurements at those scales and can eventually identify new modes of behavior within biological systems~\cite{wang2023scientific}.

This vision is possible thanks to advances in LLMs~\cite{touvron2023llama,team2023gemini,radford2018improving}, multimodal learning, and generative models.  Chat-optimized LLMs, such as GPT-4~\cite{vemprala2023chatgpt}, can incorporate feedback, enabling AI agents to cooperate through conversations with each other and with humans~\cite{AutoGPT}. These conversations can involve agents seeking human feedback and critique, and identifying gaps in their knowledge~\cite{yao2022react,shinn2023reflexion}. Then, since a single LLM can exhibit a broad range of capabilities---especially when configured with appropriate prompts and inference settings, conversations between differently configured agents can combine these capabilities in a modular manner~\cite{wu2023autogen}. LLMs have also demonstrated the ability to solve complex tasks by breaking them into subtasks~\cite{singh2022progprompt,huang2022language}. However, suppose we follow conventional approaches to foundation models such as LLMs and other large pre-trained models. In that case, we may not develop AI agents that can generate novel hypotheses because such novelty would not have been in the data used to train the model, suggesting that current foundation models alone are not sufficient for ``AI scientists''. Using LLMs as a comparison, generating novel hypotheses requires creativity and grounding in scientific knowledge, whereas generating novel text requires adherence to semantic and syntactic rules~\cite{krenn2022scientific}, so the latter approach aligns well with techniques for next-token prediction within LLMs, while the former does not.

Here, we offer a perspective that ``AI scientists'' can be realized as AI agents backed by humans, LLMs, ML models, and other tools like experimental platforms that form a compound AI system. An AI agent should be able to formulate biomedical hypotheses, critically evaluate them, characterize their uncertainty, and use that as a driver to acquire and refine its scientific knowledge bases in a way that human scientists can trust~\cite{sun2024trustllm}. AI agents should be designed to adapt to new biological insights, incorporate the latest scientific findings, and refine hypotheses based on experimental results. This adaptability ensures agents remain relevant in the face of rapidly evolving biological data~\cite{krenn2022scientific}, balancing between encoding new findings and retaining old knowledge~\cite{kotha2023understanding}.

Realizing this perspective shift, biomedical AI agents can impact areas ranging from virtual cell simulation, programmable control of phenotypes, and the design of cellular circuits to developing new therapies. Virtual cell simulation involves creating detailed models of cellular processes, where AI can predict the effects of genetic modifications or drug treatments on cell behavior. This can allow for an understanding of cellular mechanisms and generation of testable hypotheses, reducing the time and cost of traditional methods. Programmable control of phenotypes leverages AI agents to design precise genetic modifications to study gene functions. For example, CRISPR-based gene editing guided by an AI agent can activate or inhibit specific genes across large cell populations in a multi-round editing campaign. Each round involves identifying the next edit based on the user-specified target phenotype and experimental readout from the previous round. Designing cellular circuits involves using AI agents to predict the behavior of genetic components and optimize their arrangement to create circuits that perform tasks such as sensing environmental changes or producing therapeutic proteins.

Ethical considerations arise from biomedical AI agents~\cite{li2023ethics,goetz2023unreliable}.  Allowing them to make changes in environments through ML tools or calls to experimental platforms can be dangerous. Safeguards need to be in place to prevent harm~\cite{kumar2024ethics}.  Conversely, discovery workflows might include conversations between AI agents (but no interaction with environments is allowed). In that case, we need to consider the impact of such interactions on scientists and their reliance on AI agents.
Further, challenges uniquely relevant to biomedical AI agents include the lack of large experimental datasets that cover diverse use cases beyond the current focus on a handful of biomedical domains like structural biology and single-cell science. AI agents need to represent biomedical knowledge in a data-efficient manner and achieve strong generalization to new tasks with little or no additional training. Biomedical AI agents can assist with research and operations under human oversight, but their impact and challenges highlight the need for responsible implementation.

%% file: 02why.tex
\section*{Evolving use of data-driven models in biomedical research}
\label{sec:why}

Over the past several decades, data-driven models have reshaped biomedical research by developing databases, search engines, machine learning, and interactive and foundation learning models (Figure~\ref{fig:timeline}). These models have advanced modeling of proteins~\cite{sachs2005causal,rao2021msa,lin2023evolutionary,baek2021accurate,alipanahi2015predicting}, genes~\cite{theodoris2023transfer}, phenotypes~\cite{yu2016translation}, clinical outcomes~\cite{singh2002gene,shipp2002diffuse,kuenzi2020predicting}, and chemical compounds~\cite{Ren_2024,stokes2020deep} through mining of biomedical data.

\xhdr{Databases and search engines}
In biological research, databases (DBs) ~\cite{berman2000protein,10002012integrated,wishart2006drugbank} aggregate knowledge from experiments and studies, offering searchable repositories containing standardized biological data vocabularies. An example of such a database is the AlphaFold Protein Structure DB~\cite{varadi2022alphafold}, which includes more than 200 million protein structures predicted by AlphaFold~\cite{jumper2021highly}. Molecular search engines retrieve information from these databases~\cite{brin1998anatomy,altschul1990basic,gaulton2017chembl}. FoldSeek~\cite{van2023fast} retrieves protein structures from the AlphaFold DB by translating query structures into 3D interaction alphabet sequences and using pretrained substitution matrices.
Search engines are designed to retrieve information based on specific queries, lacking the ability to refine these queries through reasoning. They cannot iteratively process obtained information to refine results or customize subsequent actions. Additionally, while databases reduce the risk of misinformation through curated data,  they lack mechanisms to identify and remove irrelevant information.

Distinct from search engines, AI agents are capable of reasoning to formulate search queries and subsequently acquire information. Curated databases offer structured and factual information, aiding in reducing the risks associated with misinformation potentially generated by agent hallucinations~\cite{zhang2023siren,lala2023paperqa}. For example, the retrieval-augmented generation~\cite{lala2023paperqa} is equipped for AI agents to answer questions based on scientific literature. A notable feature of these agents is their ability to retrieve information when needed and to create and iteratively process the obtained passages.
This reflection process makes the agent controllable during inference, allowing for customization of its actions to meet task requirements beyond what is possible using search engines and database queries.

\xhdr{Machine learning models}
Beyond information retrieval, ML models excel in identifying patterns and assimilating latent knowledge to generalize predictions about novel data~\cite{krizhevsky2012imagenet,he2016deep}.
Existing machine learning models typically require specialized models for each task and do not possess the reasoning and interactive capabilities that distinguish AI agents.
An example is the  AlphaFold~\cite{jumper2021highly}, which predicts 3D protein structures with high accuracy using multi-sequence alignment with a deep learning model but is tailored for protein folding.
AI agents represent an evolution in ML models, building on the foundations of successes such as the transformer architecture~\cite{vaswani2017attention} and generative pretraining~\cite{radford2018improving}. These agents' reasoning and interactive capabilities distinguish them from ML models, which typically require specialized models for each task. Unlike traditional ML models, agents assess the evolving environment, which is valuable for modeling dynamic biological systems.

\xhdr{Interactive learning models}
Interactive learning, often referred to as active learning~\cite{Hernandez-Garcia_2023} and reinforcement learning~\cite{ouyang2022training}, represents a further advancement in ML models by incorporating exploration mechanisms and human feedback. Active learning strategies can help build models for datasets with small sample sizes when conventional ML models might be insufficient due to limited statistical power.
It selectively queries the most informative data points for labeling and optimizing the learning process, which improves how models learn with data.
Reinforcement learning involves an agent learning how to act by observing the results of past actions in an environment, mirroring the trial-and-error approach.
In biological research, interactive learning has been used for small molecule design~\cite{zhavoronkov2019deep}, protein design~\cite{Hie_Yang_2022,Lutz_2023}, drug discovery~\cite{Bailey_2023_DD_AL,Soleimany_2021}, perturbation experiment design~\cite{Zhang_2023_Causal_AL},
and cancer screening~\cite{Yala_2022}.
For instance, GENTRL~\cite{zhavoronkov2019deep} uses reinforcement learning to navigate the chemical space and identify chemical compounds that can act against biological targets. However, interactive models are predominantly designed for narrow use cases and struggle to generalize to new goals without retraining the models from scratch. Leveraging interactive learning, AI agents achieve greater autonomy in information retrieval tasks. Active learning improves training efficiency through data labeling selected to maximize model performance. However, AI agents extend beyond this data-centric approach; for example, reinforcement learning with human feedback (RLHF)~\cite{ouyang2022training} uses a ``reward model'' to train and fine-tune an LLM-based agent with direct human feedback to understand human instruction naturally.

\xhdr{AI agents}
Biomedical AI agents have advanced capabilities, including proactive information acquisition through perception modules, interaction with tools, reasoning, and engaging with and learning from their environments. Agents use external tools, such as lab equipment, and have perception modules, such as integrated visual ML tools, to receive information from the environment. Agents can incorporate search engines and ML tools and process information across data modalities via perception modules to generate hypotheses and refine them based on scientific evidence~\cite{boiko2023autonomous,bran2023chemcrow}.

%% file: 03how.tex
\section*{Types of biomedical AI agents}
\label{sec:achieve_agent}

The prevailing approach to building agents is to use LLMs, where a single LLM is programmed to perform various roles.
However, beyond LLM agents, we envision multi-agent systems for discovery workflows that combine heterogeneous agents  (Figure~\ref{fig:fig1}) consisting of ML tools, domain-specific specialized tools, and human experts.
Given that much of biomedical research is not text-based, such agents have broader applicability to biomedicine than LLM-based agents alone.

\subsection*{Large language model based AI agents}
\label{sec:llm-agents}

Programming a single LLM with diverse roles equips LLM-based agents with conversational interfaces that emulate human expertise and can access tools \cite{sumers2023cognitive,wang2023survey} (Figure~\ref{fig:achive_agent}a). The rationale behind this approach stems from pretraining an LLM to encode general knowledge, followed by in-domain fine-tuning of the LLM to encode domain-specific specialist knowledge and aligning the LLM with human users through role-playing and conversation. Instruction tuning \cite{wei2021finetuned} can be used for the former by training the LLM to follow human instruction through prompt examples, including dialogues that incorporate biological reasoning \cite{huang2023visual}. Additionally, RLHF optimizes LLM performance by selecting the most human-preferred outputs from a range of responses to specific prompts, further aligning LLMs with human roles.
Consequently, a single LLM, programmed to fulfill multiple roles, can provide a more practical and effective solution than developing specialized models. By assigning specific roles, the agents can replicate the specialized knowledge of experts across various fields, such as structural biology, genetics, and chemistry, surpassing the capabilities of querying a non-specialized LLM \cite{nori2023can} and performing tasks previously not possible \cite{park2023generative}.
Early results in clinical medicine question-answering suggest that assigning specific roles, such as clinicians, to GPT-4 \cite{nori2023can} can achieve better performance in terms of accuracy on multiple-choice benchmarks compared to using domain-specialized LLMs like BioGPT \cite{luo2022biogpt}, NYUTron \cite{jiang2023health}, and Med-PaLM \cite{singhal2022large,singhal2023towards}.

We envision three approaches for assigning roles to biological AI agents: domain-specific fine-tuning, in-context learning, and automatic generation of optimized roles. The first approach involves instruction-tuning an LLM across many biological tasks to ground the LLM in the biological domain, followed by RLHF to ensure that the tuned LLM performs tasks aligned with scientists' goals, wants, and needs. The second approach uses in-context learning of LLMs~\cite{mann2020language} to process longer contextual information provided in inputs, such as biologist-generated instructions, enabling agents to grasp the domain context for each task. This approach is supported by using textual prompts to define agent roles~\cite{park2023generative,wang2023voyager}. Both strategies require biologists to carefully gather task-specific data or craft prompts. However, as roles defined by humans may not always direct agents as intended, there has been a movement towards allowing LLM-based agents more autonomy in role specification. This paradigm shift in role definition enables agents to autonomously generate and refine role prompts, engaging in self-directed learning and role identification.
For instance, An agent's ability to evolve and tailor its prompts in reaction to user inputs has been demonstrated in~\cite{fernando2023promptbreeder}. Additionally, the application of LLM as an optimizer to enhance prompt refinement and optimization for improved performance in assigned roles has been investigated in~\cite{yang2023large}.
Through this self-referential learning framework, agents transition from task executors to entities capable of more autonomous learning.

The agent system, comprising a single LLM prompted to adopt various roles, has shown to be a valuable support tool in scientific research.
Studies suggest that agents allocated specific roles exhibit enhanced capabilities compared to either sequentially querying a single LLM or employing a single tool repetitively. A case in point is Coscientist~\cite{boiko2023autonomous}, which shows the potential of GPT-4-based agents for chemical research tasks, including optimizing reactions for palladium-catalyzed cross-couplings. Within Coscientist, GPT-4 undertakes the role of a planner, serving as a research assistant. The agent uses in-context prompts to use tools such as web and documentation search, code execution via Python API, and even symbolic lab language (SLL)~\cite{boiko2023autonomous}. To complete tasks that require access to a physical device, the planning agent starts with a prompt provided by the scientist and uses search tools to compile the documentation for the experiment. Following this, the agent generates SLL code and executes it, which entails transferring it onto the device and controlling the device.

\subsection*{Multi-agent AI systems}
\label{sec:agent-systems}

LLM-based agents implemented through autoregressive LLM approaches acquire skills such as planning and reasoning by emulating observed behaviors in training datasets. However, this mimicry-based learning results in limited agent capabilities, as they do not achieve a deep understanding of these behaviors~\cite{lecun2022path}. Consequently, a single agent often lacks the comprehensive skill set needed to complete complex tasks. A practical alternative is deploying a multi-agent AI system, wherein the task is segmented into more manageable subtasks. This approach allows individual agents to address specific subtasks efficiently, even with incomplete capabilities.
Distinct from single-LLM-based agents, a multi-agent system incorporates several agents endowed with specialized capabilities, tools, and domain-specific knowledge. For successful task execution, these agents must conform to working protocols. Such cooperative efforts equip LLMs with unique roles, specialized knowledge bases, and varied toolsets, simulating an interdisciplinary team of biology specialists. This approach is akin to the diverse expertise found across departments within a university or an institute.

In the following, we introduce five collaborative schemes for multi-agent systems.

\xhdr{Brainstorming agents (Figure~\ref{fig:achive_agent}b)}
Brainstorming research ideas with multiple agents constitutes a collaborative session to generate a broad spectrum of research concepts through the joint expertise of scientists and agents. In such sessions, agents are prompted to contribute ideas, prioritizing the volume of contributions over their initial quality to foster creativity and innovation. This method encourages the proposal of unconventional and novel ideas, allowing participants to build upon the suggestions of others to uncover new avenues of inquiry while withholding judgment or critique.
The process enables agents to apply their domain knowledge and resources to form a collective idea pool.
Each agent would provide insights and generate hypotheses based on their specialized knowledge, which the group can then integrate and refine.
For example, in a multi-agent system designed for Alzheimer's research, agents could specialize in microglia biology, neuronal degeneration, and neuroinflammation.
To explore new therapeutic targets for Alzheimer's disease, an agent specialized in microglia biology might propose investigating the role of microglial cells in synaptic pruning, while another agent focused on neuronal degeneration could suggest examining the protective effects of certain neurotrophic factors. These diverse ideas are pooled together, allowing researchers to explore a comprehensive range of potential research directions.

\xhdr{Expert consultation agents (Figure~\ref{fig:achive_agent}c)}
Expert consultation entails soliciting expertise from individuals or entities with specialized knowledge. This process involves expert agents gathering information from various sources and providing insights, solutions, decisions, or evaluations in response. Other agents or humans then refine their approaches based on this feedback.
LLMs have the potential to assist in offering scientific critiques on research manuscripts, as demonstrated in recent studies~\cite{liang2023can}.
However, LLMs lack the nuanced understanding of human reviewers and should be seen as complementary to, not a replacement for, human expertise.
Similarly, an AI agent might consult another agent specialized in a specific area to refine ideas within AI systems, mirroring the mentor-mentee dynamics found in academic environments. In another example, in addressing Alzheimer's and related dementias, diagnosing Alzheimer's based on cognitive criteria might present borderline cases. Consulting an AI agent could offer additional perspectives, determining if such cases align with Alzheimer's based on brain pathology or alternative biomarkers.

\xhdr{Research debate agents (Figure~\ref{fig:achive_agent}d)}
In a research debate, two teams of agents present contrasting perspectives on a research topic, aiming to persuade the agents of the opposing team. Agents are split into two groups, each adopting distinct roles for the debate. One group gathers evidence to fortify its position using various knowledge sources and tools, while the opposing group critiques this evidence, striving to expose or neutralize its weaknesses with superior evidence. The objective for each faction is to articulate their arguments more effectively than their rivals, engaging in a systematic discourse to defend their viewpoint and challenge the veracity of their adversaries' assertions. This methodology promotes critical thinking and bolsters effective communication as each team endeavors to construct the most compelling argument supporting their stance.

\xhdr{Round table discussion agents (Figure~\ref{fig:achive_agent}e)}
Round table discussions involve multiple agents engaging in a process that fosters the expression of diverse viewpoints to make collaborative decisions on the topics under discussion. In such sessions, agents articulate their ideas and insights, pose questions, and provide feedback on others' contributions. They then respond to these queries, refine their initial propositions based on feedback, or attempt to persuade their peers.
This method promotes equal participation among all agents, urging them to contribute their expertise and perspectives, offer constructive criticism, question underlying assumptions, and suggest amendments to improve the proposed solutions.
For instance, Reconcile~\cite{chen2023reconcile} implements a collaborative reasoning scheme among LLM agents through successive rounds of dialogue. Agents attempt to convince each other to adjust their responses and use a confidence-weighted voting mechanism to achieve a more accurate consensus than if a single LLM-based agent is used. During each discussion round, Reconcile orchestrates the interaction between agents using a `discussion prompt,' which includes grouped answers and explanations produced by each agent in the preceding round, their confidence levels, and examples of human explanations for correcting answers.

\xhdr{Self-driving lab agents (Figure~\ref{fig:achive_agent}f)}
The self-driving laboratory is a multi-agent system where the end-to-end discovery workflow is iteratively optimized under the broad direction of scientists but without requiring step-by-step human oversight~\cite{sanders2023biological}. Once the agent system is trained, it can describe experiments necessary to test the generated hypotheses, analyze the results of said experiments, and use them to improve its internal scientific knowledge models. Agents in the self-driving system need to address the following three elements: determine inductive biases to reduce the search space of hypotheses, implement methods to rank order hypotheses considering their potential biomedical value with experimental cost, characterize skepticism via uncertainty quantification and analysis of experiments in reference to the original hypothesis, and refine hypotheses using data and counterexamples from experiments~\cite{davies2021advancing}. Ideally, hypothesis agents are creative and reflective when developing biological hypotheses that extrapolate indirectly from the existing body of knowledge~\cite{krenn2022scientific}.
There is emerging evidence that generative models have the potential to generate novel hypotheses. \cite{tshitoyan2019unsupervised} demonstrated that using latent knowledge from published materials science literature can recommend novel materials. \cite{jablonka2024leveraging} leveraged LLMs trained with an autoregressive pretraining objective to predict molecules.
Experimental agents steer operational agents that use a combination of in silico approaches and physical platforms to execute experiments. Reasoning agents integrate the latest results to guide future experimental design. The utility of experimental results, such as the yield of high-throughput screening of a chemical library against a biological target, can be compared for different versions of the agent system given a time budget for hypothesis and experiment generation.

%% file: 04levels.tex
\section*{Levels of autonomy in AI agents}
\label{sec:agent_level}

When integrated with experimental platforms, AI agents can operate at varying levels of autonomy tailored to the diverse requirements across biomedical fields. We classify these AI agents into four levels according to their proficiency in three areas of discovery: Hypothesis, Experiment, and Reasoning (Table~\ref{tab:levels_table}). Specific capabilities within each area define these levels, necessitating that agents exhibit the capabilities for a given level across all areas (an agent with Level 3 capabilities in the Experiment area but only Level 2 capabilities in Reasoning and Hypothesis areas would be classified as Level 2).

Level 0, denoted as \textbf{no AI agent}, uses ML models as tools coordinated by interactive and foundation learning models.
At this level, ML models do not independently formulate testable and falsifiable statements \cite{glass2008brief} as hypotheses. Instead, model outputs help scientists to form precise hypotheses. For example, a study employed AlphaFold-Multimer to predict interactions of DONSON, a protein with limited understanding, leading to a hypothesis about its functions~\cite{lim2023silico}. Level~1, termed \textbf{AI agent as a research assistant}, features scientists setting hypotheses, specifying necessary tasks to achieve objectives, and assigning specific functions to agents. These agents work with a restricted range of tools and multimodal data to execute these tasks. For instance, ChemCrow \cite{bran2023chemcrow} combines chain-of-thought reasoning~\cite{wei2022chain} with ML tools to support tasks in organic chemistry, identifying and summarizing literature to inform experiments. In another example, AutoBa \cite{zhou2023automated} automates multi-omic analyses. These two agents are designed for narrow scientific domains; ChemCrow and AutoBa optimize and execute actions to complete tasks that are designed and predefined by scientists. Level 1 agents \cite{tang2023medagents, hu2023novo, zhou2023automated, bran2023chemcrow} formulate simple hypotheses inferred from existing knowledge and utilize a limited set of tools, lacking the capacity necessary to achieve Level 2 autonomy.

At Level 2, \textbf{AI agent as a collaborator}, the role of AI expands as scientists and agents collaboratively refine hypotheses. Agents undertake tasks critical for hypothesis testing, using a wider array of ML and experimental tools for scientific discovery.
\cite{tshitoyan2019unsupervised}
However, their capability to understand scientific phenomena and generate innovative hypotheses remains constrained, highlighting a linear progression from existing studies. The transition to Level 3, or \textbf{AI agent as a scientist}, marks a major evolution, with agents capable of developing and extrapolating hypotheses beyond the scope of prior research, synthesizing concepts beyond summarizing findings and establishing concise, informative, and clear conceptual links between findings that cannot be inferred from literature alone, eventually yielding a new scientific understanding. While multiple Level 1 agents exist across various scientific fields, Levels 2 and 3 agents have yet to be realized.

The levels of autonomy described for artificial general intelligence (AGI) agents in scientific contexts, particularly in biology, deviate from existing taxonomies that focus on general human-AI interaction separate from the collaborative dynamics between scientists and AI. Existing taxonomies of autonomy consider solely the balance of responsibilities between AI agents and humans---with no consideration of biomedical discovery---and focus on developing AGI to surpass human performance across varying skill levels \cite{morris2023levels}.

As the level of autonomy increases, so does the potential for misuse and the risk of scientists developing an overreliance on agents. While agents have the potential to enhance scientific integrity, there are concerns regarding their use in identifying hazardous substances or controlled substances \cite{Urbina_2022}.
Responsible development of agents requires developing preventive measures \cite{tang2024prioritizing, baker2024protein}.
The responsible deployment of agents must account for the risk of overreliance, particularly in light of evidence that LLMs can produce convincing but misleading claims and spread misinformation. The risks will likely increase as agents undertake more autonomous research activities. Agents must be scrutinized as scientists, including reproducibility and rigorous peer review of agentic research.
We illustrate these definitions of levels by giving examples of progression between the levels in genetics, cell biology, and chemical biology. We selected these areas because of the availability of large datasets that have recently driven the development and application of ML models. We describe the challenges and limitations that biomedical AI agents may present for an enhanced understanding and progression through levels of autonomy.

\subsection*{Illustration of AI agents in genetics}

Research in human genetics seeks to understand the impact of DNA sequence variation on human traits. LLM-based agents operating at Level 1 would perform specific tasks relevant to genetic studies. For instance, in a genome-wide association study (GWAS), a Level 1 agent can write bioinformatics code to process genotype data to (1) execute quality control measures, such as the removal of single-nucleotide polymorphisms (SNPs) missing in many individuals or control for population stratification \cite{Marees2018-ga}, (2) estimate ungenotyped SNPs through imputation, and (3) conduct the appropriate statistical analyses to identify relevant SNPs, taking into account the false discovery rate \cite{Uffelmann2021}. Following the analysis, the Level 1 agent reviews and reports findings, including any filtered SNPs and rationales for their exclusion.

Instead of executing narrow tasks following human instruction, a Level 2 agent identifies and executes tasks independently to refine a hypothesis initially given by the scientist. For example, it may explore the effectiveness of drugs for a patient subgroup within complex diseases, where genetic underpinnings can influence drug response \cite{Frueh2010-xu}. Given a hypothesis that a particular drug is effective in a subset of patients with idiopathic or genetic generalized epilepsy (GGE)—a condition with a robust genetic causality \cite{panayiotopoulos2005epilepsies}—a Level 2 agent would synthesize genetic information from GWAS meta-analyses \cite{epilepsy-gwas-meta-analysis}, such as the UK Biobank \cite{ukbiobank}, targeted sequencing studies \cite{Gamirova2024}, and knowledge bases like Genes4Epilepsy \cite{Oliver2023-ya}. The agent identifies GGE subtypes and causal genes by analyzing patient genetic data, predicting which subgroups might benefit from the drug based on genetic markers. It would then conduct in vitro functional studies to confirm these predictions, ultimately presenting evidence on how the drug could benefit GGE patient subpopulations by synthesizing concepts beyond summarizing findings.

Level 3 agents coordinate a system of agents (Figure~\ref{fig:achive_agent}) to discover and evaluate gene markers for specific phenotypes. These agents help initiate new study groups and optimize non-invasive methods of DNA collection for cost-effectiveness and recruitment processes \cite{Salowe2022-fo}. Once data are collected, the agents innovate statistical methods to identify causal variants from genotypic data amidst confounders such as linkage disequilibrium \cite{Aissani2014} and develop in vitro techniques for validating candidate gene markers in disease models. Level 3 agents collaborate with scientists to generate and test hypotheses for comprehensive genetic insights.

\subsection*{Illustration of AI agents in cell biology}

Cells are fundamental units of study in cell biology. Advances in single-cell omics, super-resolution microscopy, and gene editing have generated datasets on normal and perturbed cells, covering areas such as multi-omics~\cite{Regev_2017_hca, Subramanian_2017_cmap, Mitchell_2023_prot_moa}, cell viability~\cite{Ghandi_2019_depmap}, morphology~\cite{Chandrasekaran_2022_jump_cp}, cryo-electron microscopy and tomography~\cite{DeTeresa-Trueba_2023_cryoet, Schiøtz_2023_cryoet}, and multiplexed spatial proteomics~\cite{Lundberg_2019_spatial_prot, Cho_2022_opencell}. This proliferation of data has spurred interest in in silico cell modeling~\citep{Johnson_2023_virtal_cell}.

ML tools have been instrumental in analyzing data across these cellular modalities, but as Level 0 agents, they lack autonomous research capabilities.
At Level 1, agents integrate specialized Level 0 models to assist in hypothesis testing. These agents actively assist scientists in developing hypotheses by synthesizing literature and predicting cellular responses using integrated models.
For example, to help investigate the resistance mechanism of a compound, Level 1 agents predict its effects in various cellular contexts~\cite{li2023contextualizing}.
These predictions also inform experimental design, such as spatial transcriptomic~\cite{Russell_2024_slidetag} and proteomic~\cite{Wik_2021_pea,Liu_2023_spatial_cite_seq} screening. Agents then retrieve and refine experimental protocols for execution on platforms~\cite{Yoshikawa_2023} and apply predefined bioinformatics pipelines, as instructed by scientists.

Level 2 agents execute predefined tasks and generate hypotheses on cellular functions and responses. They autonomously define and refine tasks to support scientific reasoning, enabling practical exploration of complex phenotypes like drug resistance.
By managing the experimental cycle and continuously updating their in silico tools, Level 2 agents actively optimize experiments to focus on key variables of resistance based on a synthesis of predictive content, uncertainty, and newly acquired data, with iterative feedback from scientists~\cite{Zhang_2023_Causal_AL}.
Level 2 agents thus form a prototype for a virtual cell model capable of hypothesis generation, encompassing closed-loop integration of digital and experimental platforms.

Level 3 agents respond to existing challenges and anticipate future directions in cell biology research.
They form hybrid virtual cell models, an organic combination of AI tools (digital agents) with high-throughput platforms (experimental agents).
Digital agents, such as LLM-based agents, autonomously identify critical knowledge gaps through literature synthesis based on criteria such as data volume, biological relevance, and clinical needs and simulate any perturbagen (extrinsic events such as gene knockouts and overexpression, compounds, cell-cell interactions; intrinsic events such as cell cycle) in any context.
Experimental agents not only optimize experimental protocols~\cite{Dixit_2016_perturbseq,Chandrasekaran_2022_jump_cp,Binan_2023} to enable high-throughput multimodal measurements but also develop transformative technologies to enable probing at unprecedented resolution across space and time across in vitro, ex vivo, and in vivo models, uncovering pioneering insights.
The ability of level 3 agents to drive the discovery of novel biological mechanisms and therapeutic strategies shifts the role of scientists from performing operational tasks to ideation and managing hybrid cell models.

\subsection*{Illustration of AI agents in chemical biology}
A major focus for chemical biology is understanding molecular interactions within cells to manipulate biological systems at molecular and cellular levels. An AI agent could analyze any molecular interaction, help design new drugs, and provide more valuable chemical probes for biological systems.

Despite considerable advances in applying ML to chemical biology, current approaches fall in Level 0. Scientists oversee all activities by integrating ML tools for structure prediction, docking, chemical synthesis, and molecular generation. At Level 1 the agent has elementary reasoning of chemical biology and can execute simple tasks autonomously such as running ML tools, or designing experiments for a given objective. However, due to limited reasoning capabilities, the agent may fail to explain more complex concepts, such as how the dynamics of molecules may influence the effects of drugs on binders or explore novel molecular scaffolds. For a level 2, the long-term objective is its function as a collaborator for scientists through excelling at tasks that are explicit continuations of existing scientific research, such as improving the efficiency of chemical probes, autonomously designing and testing de novo enzymes, or designing new binders by leveraging trends in related targets. Level 2 AI agents have deeper expertise in more domains, such as retro-synthesis, crystallography, bioassays, and directing robotic arms to conduct research.

The goal of a Level 3 agent in chemical biology is the ability to study all types of molecular interactions in a cell. This agent would work alongside human scientists to explore research questions that are challenging for the field, such as binder design for undruggable targets \cite{dang2017drugging}, significantly improving specificity and efficiency of in vivo bioorthogonal reactions, or developing new chemical probes that can access new spatial and temporal scales. Unlike the Level 2 agent's use of well-established protocols, a Level 3 agent aims to unlock experimental capabilities that are not currently accessible. For example, AI agents could be tasked to probe molecular dynamics at longer timescales than what is currently accessible. At this level, agents have a thorough understanding of existing literature and work alongside scientists to unlock new fields of chemical biology.

%% file: 05components.tex
\section*{Roadmap for building AI agents}
\label{sec:roadmap}

An AI agent is built as a compound system that consists of modules~\cite{xi2023rise,sumers2023cognitive,wang2023survey} each implementing a distinct functionality.
Here, we describe these modules (Figure~\ref{fig:componenet_agent}), focusing on perception, interaction, memory, and reasoning modules necessary for AI agents to interact with humans and engage with experimental environments. Interactions between the agent and its environment are characterized by two elements: the agent's perception of its surroundings and its subsequent engagement with them. Perception modules enable the agent to interpret and assimilate information from various data modalities. Then, learning and memory allow agents to interact with an environment and complete tasks, by acquiring new knowledge and retrieving previously learned one. Finally, the reasoning module processes information and executes action plans. Using a published study as an example~\cite{lieber2019mitochondrial},  Figure 5e illustrates a hypothetical AI agent that examines the molecular mechanisms of selective removal of mitochondrial DNA mutants in the Drosophila female germline through perception, interaction, memory, and reasoning modules.

The division of research into smaller tasks handled by AI agents presents an intriguing approach, building on the success of modular and sequential bioinformatics workflows like Snakemake and Docker. Unlike these workflows, which are often static and require manual updates and reconfiguration to handle new tasks or integrate new tools, AI agents are dynamic and operate in a personalized, user-specific, and context-appropriate manner. They can learn to use new tools and adjust their workflows based on the specific instructions and needs of the scientist. 
Further, the adaptive allocation of tasks by AI agents can be helpful in automatically incorporating new tools and restructuring existing pipelines, much like a human researcher would. For example, AI agents could experiment with and create new protocols beyond the currently established methods in integrating multimodal omics data. For instance, while established protocols for integrating multi-modal, such as scRNA-seq with scATAC-seq or spatial data, exist, AI agents could develop new pipelines for multi-modal integrations beyond the three modalities, or multi-scale integrations such as atlas-scale single-cell and bulk RNA-seq data, or normal and disease state data from cell lines, organoids, and patient samples, based on their initial attempts. 

\subsection*{Perception modules}

Perception modules equip LLM-based agents with the capability to understand and interact with elements in the environment in which they operate, such as biological workflows and human users. For perception, agents need to integrate abilities to receive feedback from multiple sources: scientists~\cite{ouyang2022training}, the environment~\cite{park2023generative}, and other AI agents~\cite{li2023camel,wu2023autogen}. This requires accommodating a diverse array of modalities. These include text descriptions~\cite{touvron2023llama}; images from light and (cryo-)electron microscopy to assess cellular processes across many conditions simultaneously~\cite{DeTeresa-Trueba_2023_cryoet, Schiøtz_2023_cryoet,liu2023llava}; videos from live imaging to assess developmental processes or animal behaviors across time~\cite{Chen_2020_NN}; longitudinal biosensor readouts and genomics profiles of cells~\cite{driess2023palme};
mass spectrometry-based proteomics to decipher protein homeostasis~\cite{li2021proteome,lin2023evolutionary}; and miniaturized platforms for conducting biochemical assays and 3D culture systems that mimic the physiological context of organ systems~\cite{Yoshikawa_2023}. 

AI agents can take different approaches to interacting with environments. The most direct one involves using natural language, which represents a common perception modality for LLM-based agents. Other techniques involve multi-modal perception modules, where agents process multi-modal data streams from the environment or align multi-modal inputs with text-based LLMs.

\xhdr{Conversational modules} 
With the rise of ChatGPT, the ability of AI agents to interpret natural language has reached such a high level~\cite{ouyang2022training} that it is now possible to build interfaces to agent systems that are entirely based on natural language with limited misinterpretations. The main focus is chat interfaces that preserve conversational history in a scrolling window, where users can converse with agents in a manner that resembles the standard approach of written human-to-human interaction. 
This approach allows scientists to express their queries using their language, promoting initiative and enabling them to precisely describe what they want. We envision that agents will maintain a history of interaction with scientists using natural language, which, in turn, will allow us to keep track of scientific interactions with agents ~\cite{park2023generative,wang2023voyager}. Combining the history trace of these interactions with retrieval-augmented generation (RAG), it will be possible to develop personalized discovery workflows tailored to individual scientists. 

\xhdr{Multi-modal perception modules}
Agents align LLMs with other data types to consider data modalities beyond natural language. This approach helps agents better model the changing environment in which the agent acts and dynamically adjust its outputs to new situations, such as evolved biological states in a virtual cell model. The alignment process involves two main strategies: textual translation and representation alignment. Textual translation converts inputs into a textual format, such as transforming data from robotics into textual descriptions that log environmental states~\cite{vemprala2023chatgpt}. 
For example, when handling readouts from experimental devices, the readouts can be combined with a textual description of their meaning, allowing the LLM to understand the readouts as a new modality. 
Alternatively, through representation alignment, data from different modalities are analyzed by modality-specific models to generate representations, such as using the visual encoder from CLIP~\cite{radford2021learning} for visual information processing. These representations are then aligned with LLM textual representations through instruction tuning~\cite{zhu2023minigpt,liu2023llava}, enabling agents powered by LLMs to perceive and interpret multi-modal data. 
For instance, to make LLM-based agents handle the protein structure data, an additional encoder is required to encode the protein structure data into a representation aligned with the LLMs’ representation space. This encoder is pre-trained with modality-specific training schemes, and an adaptor is placed between this encoder and LLMs to align the representations of the two modalities. Then, instruction tuning is applied using data containing both modalities to train the adaptor for alignment.
An alternative to alignment involves allowing the agents to receive input expressed in different modalities~\cite{team2023gemini,fuyu-8b}. For instance, Fuyu~\cite{fuyu-8b} uses a decoder-only transformer architecture to process image patches and text tokens jointly. Similarly, Gemini~\cite{team2023gemini} is engineered to handle visual, audio, and text inputs within a single model. Once perception modules are implemented for agents to receive inputs from the environment, modules for interaction and reasoning follow to process the inputs and interact externally. 
Training agents with strong perception abilities on biomedical data requires extensive, high-quality data pairs that align multiple modalities. However, collecting such data remains challenging. For example, multimodal experimental platforms are non-existent or have low-throughput yields, certain tissues, and cell types are not experimentally available, and a long tail of disease phenotypes has small sample sizes, making data collection infeasible.

\subsection*{Interaction modules}
\label{sec:interact}

Beyond conversational modules, scientists use ML-based and other tools in biological research, explore datasets through graphical user interfaces (GUIs) to analyze and visualize data, and engage with physical equipment and wet lab experimental platforms. Chat-optimized LLM-based agents thus need interaction capabilities to communicate and collaborate with scientists, other AI agents, and tools to function beyond a simple chatbot. Agents must incorporate essential interaction modules to interact with elements in the environment. These include agent-human interaction to support communication with scientists and following human instruction~\cite{kopp2021revisiting,huang2022inner}, multi-agent interaction for collaboration among agents, and tool-use action to access ML tools and experimental platforms.

These interaction abilities of LLMs, when combined with interactive `function calling' (i.e., LLM requesting for tasks to be completed), can act as an intermediary between scientists and the agent's interface, as well as between scientists and various functional items, such as tools and other agents. This approach allows scientists to express their intentions in natural language without needing to search for how and where to accomplish tasks. At the same time, the advantages of functional items are preserved because agents can interact with tools and use them to provide feedback. However, interactive modules trained on general, non-biological domains might not be well-suited for specialized biomedical terminologies, requiring in-domain training on biomedical tools.

\xhdr{Agent-human interaction modules}
The interaction between scientists and AI agents synchronizes scientific objectives with AI agents through cooperative communication and modeling of biological knowledge. Natural language processing and human evaluation methods are predominantly used to develop this interaction capability. InstructGPT~\cite{ouyang2022training} enhances the GPT model through supervised fine-tuning with examples of human dialogues to improve the model's conversational skills. The alignment between agents and humans can be refined through RLHF, which adjusts the model based on a reward model trained using human assessments of the model's responses. Alternatively, RLHF can be replaced by direct preference optimization~\cite{rafailov2023direct}, which is a parameterized method that provides a more consistent and efficient alignment with human preferences.
Through agent-human interaction, agents become attuned to human needs and preferences~\cite{huang2022inner,AutoGPT}, using human insight as a directive for carrying out complex tasks~\cite{wu2023autogen}. For instance, Inner Monologue~\cite{huang2022inner} employs human feedback to discern user preferences or interpret ambiguous requests in an embodied context. In AutoGPT~\cite{AutoGPT}, humans formulate tasks and score solutions returned by agents, and AutoGen~\cite{wu2023autogen} can use human expertise to solve tasks better than agents alone.

\xhdr{Multi-agent interaction}
Multi-agent interactions support solving complex goals that agents could not complete if they operated independently. In such interdisciplinary systems, agents that could specialize in different biological domains, each with distinct capabilities, engage in interactions through various communication means. 
Language has emerged as the predominant medium for multi-agent interactions due to the ability of agents to communicate with humans linguistically~\cite{guo2024large,nascimento2023self,wu2023autogen,li2023camel,chen2023reconcile}. An instance of this is Generative agents~\cite{park2023generative}, which create interactive environments where agents mimic human behavior and interact using natural language. Different strategies are used for multi-agent interaction, including cooperation~\cite{lowe2017multi,hong2023metagpt,zhang2023building} and negotiation~\cite{chen2023reconcile,liang2023encouraging,fu2023improving}. For example, MetaGPT~\cite{hong2023metagpt} applies standardized operating procedures from human teamwork to define tasks and agent responsibilities. 

Through these approaches, agent interactions make it possible to tackle tasks that are too complex for just one agent to handle~\cite{mandi2023roco,tang2023medagents}. MedAgent~\cite{tang2023medagents} leverages the expertise of multiple medical AI agents for medical reasoning. Similarly, RoCo~\cite{mandi2023roco} employs robot agents with varied roles to accomplish complex tasks in the physical world. Multi-agent interaction can also boost the proficiency of less skilled agents by allowing them to learn from more experienced counterparts~\cite{saha2023can}. These interactions also enable the creation of simulations for a variety of environments, ranging from public health scenarios~\cite{williams2023epidemic} to human social behaviors~\cite{park2023generative,park2022social}, enhancing the system's adaptability and application in diverse contexts.

\xhdr{Tool use}
To manage tasks from diverse environments, agents require tools to boost their capabilities~\cite{parisi2022talm}. 
Commonly used tools are application APIs~\cite{schick2023toolformer}, search engines~\cite{nakano2021webgpt}, ML models~\cite{shen2023hugginggpt}, knowledge databases~\cite{hu2023chatdb}, and robotic machinery for physical tasks~\cite{coley2019robotic,ahn2022can,vemprala2023chatgpt}. 
Different Level 1 agent systems have been developed that can interact with one or more types of tools. 
ChemCrow~\cite{bran2023chemcrow} leverages chemical tools and search engines to address chemical challenges. 
WebGPT~\cite{nakano2021webgpt} can conduct searches and navigate web browsing environments. 
SayCan~\cite{ahn2022can} controls a robot in the physical world using an LLM to complete tasks.
To invoke these tools, AI agents generate commands in specific formats~\cite{shen2023hugginggpt,schick2023toolformer,hu2023chatdb} or query pre-trained control models to execute actions~\cite{ahn2022can,ramesh2021zero}. To develop these capabilities, agents can use in-context learning~\cite{shen2023hugginggpt} or fine-tuning with tool-use demonstrations~\cite{schick2023toolformer}, where the latter represents a more sophisticated approach. 

In the case of in-context learning, it is necessary to include system abilities in the prompt so agent systems can use `function calling' to query tools.
For example, HuggingGPT~\cite{shen2023hugginggpt} uses ChatGPT as a controller to integrate all ML models on Hugging Face through in-context learning.
The alternative approach consists of using model fine-tuning with `function calling' to create an LLM-based agent with integrated abilities of a function/tool. For instance,
Toolformer~\cite{schick2023toolformer} introduces a self-supervised learning method to master the use of tools' APIs with minimal demonstrations for each API. \par

By modeling scientists' needs by analyzing natural language textual inputs, AI agents can select the most likely available tool, identify the desired user interface component, and execute the scientist's expected actions. Interaction modules are designed to be integrated and adapted to suit changing environments. For Level 2 and Level 3 agents, agents autonomously learn new types of interactions and how/when to start using new tools.

\subsection*{Memory and learning modules}

When using tools and ML models for biological research,  
scientists keep records of experimental logs and plan their next steps based on them.
In AI agents, memory modules alleviate the need for manual log recording by memorizing necessary experimental outputs.
Contrary to ML models that perform one-time inference to generate predictions, memory modules in LLM-based agents store and recall information. This is necessary for executing complex tasks and adapting to new or evolving environments. Memory modules are designed to store long-term and short-term learned knowledge. As agents encounter new situations and acquire data, memory modules get updated with new information.

\xhdr{Long-term memory modules}
Long-term memory stores essential and factual knowledge that underpins agent behavior and understanding of the world, ensuring this information persists beyond task completion. This memory can be internal, encoded within the model's weights via learning processes~\cite{radford2018improving,hu2021lora}, or external, maintained in auxiliary knowledge bases~\cite{qian2023communicative,zhou2023llm}. Internal memory is directly used for accomplishing zero-shot tasks~\cite{touvron2023llama,team2023gemini} while accessing external memory requires actions by the agent to fetch and integrate data into short-term memory for immediate use~\cite{zhu2023ghost,neelakantan2022text}. For instance, ChatDB~\cite{hu2023chatdb} uses an external database for memory storage, and MemoryBank~\cite{zhong2023memorybank} encodes memory segments into embeddings for later retrieval. Agents can query knowledge banks, such as a GWAS database to find genetic evidence for a candidate protein target, a knowledge base of therapeutic mechanisms of action, and scientific literature with up-to-date information for the agent to integrate and decide whether the protein can be modulated through a therapeutic perturbation (Figure~\ref{fig:bio_agent_example}b).
The learning process updates long-term memory by adding new knowledge or replacing outdated information. Internal memory of an agent can be updated using parameter-efficient fine-tuning~\cite{hu2021lora,dettmers2023qlora}, interactive learning~\cite{ouyang2022training}, and model editing~\cite{meng2022locating}. These strategies must be effective for large models~\cite{dettmers2023qlora} and avoid the loss of previously learned information~\cite{zhang2024comprehensive}. On the other hand, updating external memory is more straightforward, involving modifications to the knowledge base~\cite{hu2023chatdb,zhong2023memorybank}. For example, in drug discovery, updating long-term memory by adding a new compound in development to the drug bank is a convenient way to maintain an up-to-date agent.

\xhdr{Short-term memory modules}
AI agents use short-term memory to temporarily store information during their interactions. This short-term memory is enabled through in-context learning, where relevant information is integrated as context prompts~\cite{rana2023sayplan,ahn2022can} or via latent embeddings~\cite{zhu2023minigpt,liu2023llava} in LLMs. For chatbots, previous conversations are kept as text prompts, supporting multiple rounds of dialogue~\cite{ouyang2022training,chiang2023vicuna}. The text-based approach lays the groundwork for communication in multi-agent~\cite{chen2023reconcile,fu2023improving} and agent-human scenarios~\cite{AutoGPT,wu2023autogen}.
In embodied AI agents, environmental feedback~\cite{ahn2022can,rana2023sayplan} is captured in textual format, acting as a short-term memory that aids reasoning. Following perception, multi-modal inputs are converted into latent embeddings, which function as short-term memory. LLaVA~\cite{liu2023llava} uses latent embeddings generated by visual encoders to retain visual information.
Short-term memory allows agents to temporarily acquire skills, such as tool usage~\cite{shen2023hugginggpt,schick2023toolformer}, store information about recent states of a biological system~\cite{chiang2023vicuna,rana2023sayplan}, and keep track of outcomes from earlier reasoning efforts~\cite{yao2022react}. This learning mechanism is crucial for agents to learn and apply new knowledge under new conditions. Moreover, short-term memory can temporarily override long-term memory, allowing agents to precede recent information over older knowledge within their model weights~\cite{li2022large}. Agents can be informed by past experiences stored in their short-term memory to tell which experiments to run in the future. In Figure~\ref{fig:bio_agent_example}a, we detail an example where the agent recalls experiments for a homologous protein to inform the initial inhibitor design for the given protein.

\subsection*{Reasoning modules}
\label{sec:reasoning}

Biological research involves a multidisciplinary and multistage process that integrates the expertise of scientists from various disciplines. Scientists formulate hypotheses, design experiments based on these hypotheses, interpret the results, and plan the next steps. 
The integration of reasoning capabilities in AI agents can assist biological research throughout this process. 
Reasoning improves agents' capabilities to plan experiments, make decisions on biological hypotheses, and resolve competing candidate biological mechanisms. 
AI agents that use large language models can implement interactive dialogue systems to explain ML models through natural language conversations.
Reasoning modules can be implemented using prompting~\cite{kojima2022large} and few-shot in-context learning~\cite{wei2022chain}. Additionally, agents can use planner models~\cite{liu2023llm,dagan2023dynamic} and action models~\cite{rana2023sayplan}. We classify reasoning modules into two categories: direct reasoning and reasoning with feedback, depending on whether agents adjust their plan in response to experimental or human feedback. 

\xhdr{Direct reasoning modules}
In direct reasoning, an agent performs planning and reasoning based on the current state of the environment, which can follow different reasoning patterns, such as single-path and multi-path reasoning. Single-path reasoning involves the agent breaking down the task into multiple recursive steps~\cite{zhang2023igniting}. For instance, chain-of-thought (CoT) reasoning allows agents to reason step-by-step either by using in-context examples~\cite{wei2022chain} or by applying a zero-shot prompt like "Let's think step-by-step''~\cite{kojima2022large}. Leap-of-thought~\cite{zhong2023let} encourages the model to use creative rather than logical reasoning. Although single-path reasoning matches well with certain situations~\cite{raman2022planning}, its ability to adjust to different conditions is limited.

Conversely, multi-path reasoning examines several paths before consolidating them into a final plan~\cite{yao2023tree,wang2023recmind}, allowing for a more thorough planning process that accounts for different scenarios. For example, Least-to-Most prompting~\cite{zhou2022least} breaks down tasks into subproblems solved sequentially. Self-consistent CoT~\cite{wang2022self} chooses the most consistent answer from a set of CoT answers. Tree-of-thoughts~\cite{yao2023tree} extends reasoning paths into a tree-like structure, generating multiple paths from each thought node and using search algorithms to select the final path. Graph-of-thoughts~\cite{besta2023graph} further develops reasoning paths into a graph structure for complex reasoning. To identify the optimal path, methods such as voting strategies~\cite{wang2022self}, Monte Carlo tree search~\cite{hao2023reasoning}, and breadth/depth-first search algorithms~\cite{yao2023tree} are used. 
Through direct reasoning, agents can generate multiple threads of thought that could consider the best pathways, protein targets, and experiments that can be run to test the role of a candidate protein target (Figure~\ref{fig:bio_agent_example}c).

\xhdr{Reasoning with feedback}
Experimental and human feedback can help AI agents to improve reasoning and planning processes~\cite {yao2022react,wang2023voyager,zhu2023ghost}. This feedback may include agent-human interaction and responses from agents, which can be complementary biological assays quantifying downstream effects of target molecules~\cite{madaan2023self}. 
In each reasoning cycle, React~\cite{yao2022react} incorporates insights from previous actions to refine its thought process and inform future actions. 
LLM-Planner~\cite{song2023llm} dynamically adjusts plans based on new observations in an embodied environment. 
Inner Monologue~\cite{huang2022inner} uses both passive and active scene descriptions and feedback from recent actions to guide future actions. 
Voyager~\cite{wang2023voyager} improves planning for subsequent steps by considering environment feedback, execution errors, and self-verification.

Beyond external feedback, an agent's feedback mechanism enables self-assessing the initial plan~\cite{madaan2023self,chen2023teaching}. Techniques like self-refine~\cite{madaan2023self} revise action outputs based on the LLM evaluation, the self-check~\cite{madaan2023self} mechanism allows the agent to review and adjust its reasoning, and reflection~\cite{shinn2023reflexion} mechanisms use prompt agents to update their decision-making. These techniques incorporate feedback from biologists, such as exploring experimental methods and environmental constraints like lab inventory (Figure~\ref{fig:bio_agent_example}d). Reasoning capabilities are necessary for generating hypotheses and conducting experiments. Generating novel hypotheses requires modeling general biomedical knowledge, the specific information on the current state of a biological system, and consideration of potential next steps. LLM-based agents can generate hypotheses through in-context reasoning, but careful selection is necessary to ensure high-quality hypotheses~\cite{wang2023hypothesis}.

%% file: 06challenges.tex
\section*{Challenges}
\label{sec:challenges}

The perspective outlines key steps to implement AI agents in biomedical research and highlights areas that can benefit from agentic AI. Challenges remain and may, in some cases, be amplified when multi-agent systems become available (Figure~\ref{fig:challenges}).

\subsection*{Robustness and reliability}

A barrier facing the deployment of agent systems -- specifically those categorized within Levels 2 and 3 as discussed inTable~\ref{tab:levels_table} -- is their propensity for generating unreliable predictions, including the hallucination of non-factual information, reasoning errors, systematic biases, and failures in planning when connected with tools and experimental platforms. These issues can be exacerbated by overconfidence in such flawed predictions (agents lack awareness of their knowledge gaps) and high sensitivity to the precise formulation of queries, particularly in the context of LLM-based agents. This behavior has been traced to how these models are trained. 
In particular, autoregressive loss compares the predicted word sequence with the actual sequence in the training data. The performance of a model trained with this loss is determined by three factors: the probability distribution of the inputs, the sequence of generated outputs, and the frequency of different tasks encountered during training~\citep{mccoy2023embers}.
As a result, model performance degrades on task variants that deviate from the assumptions made during training \citep{wu2023reasoning}.

Sensitivity to input and task probability also offers a potential explanation for the widely observed success of various prompting techniques \citep{wei2022chain, nye2021show, yao2023tree} (methods to paraphrase the same query). By providing informative context, instructive reasoning steps, or representative examples, these techniques can act as an empirical means by which task and input probability (and, thus, model performance) are increased. However, crafting high-quality prompts tends to be highly empirical while requiring significant effort and domain knowledge.

Beyond the linguistic domain, even the most advanced models fail in tasks with real-world entities that require physically meaningful actions, posing an obstacle to embodied agents. While embedding continuous sensor data into a language model can lead to improvements \citep{driess2023palme}, limitations to understanding physical interactions and long-horizon planning remain. The complexities of training such multi-modal systems, the need for large datasets to cover the range of embodied tasks and environments, and the computational demands of processing multi-modal inputs all remain open questions \citep{team2023gemini}. Deployment faces challenges from false negatives causing repeated attempts and eventual stalling of the embodied agent \citep{huang2022inner}. Hence, it is necessary to verify the agent action plan before execution.

Uncertainty quantification can trigger fall-back safety measures like early termination, pre-defined safe maneuvers, or human-in-the-loop interventions. However, foundation models cannot reason about the uncertainty associated with their outputs, and no well-established statistical protocol exists for increasingly ubiquitous architectures \citep{chen2023quantifying, vaswani2017attention}. Techniques such as various forms of prompting, \eg, \citep{tian2023just, wang2022self, kuhn2023semantic} estimate uncertainty based on the model's predictive distribution, p(output|input), which may itself be subject to bias (\citep{mccoy2023embers}, Section 3.3); furthermore, it does not consider the distribution of model parameters consistent with the observed training data and marginalizes over its predictions \citep{shafer2008tutorial}. While conformal prediction \citep{JMLR:v9:shafer08a} has emerged as a framework for uncertainty estimation of model predictions, its sensitivity to the choice of underlying statistical assumptions and the calibration of confidence levels have been criticized. The lack of a default technique is partly due to the difficulty of establishing a thorough quality assessment of uncertainty estimates. This makes it difficult to make choices in agent design and to reassure users about its calibration.

One concern is that advanced capabilities come at the cost of compromised transparency and the risk of misalignment. For instance, integrating human feedback can promote desirable agent behavior, but it can also exacerbate persuasive abilities, echoing false beliefs \citep{perez-etal-2023-discovering}. Fine-tuning existing models with new data can compromise their original alignment, challenging the integrity of the AI agent's intended purpose \citep{qi2023finetuning}. Jailbreak attacks can similarly affect post-deployment, highlighting the need for rigorous evaluation \citep{wei2023jailbroken}. 

Errors are inevitable in complex multi-agent systems, making their management crucial to maintaining system robustness and reliability. Due to their interactive nature, these systems are sensitive to compounding errors, where small issues can escalate into significant problems if not addressed promptly. Effective error management strategies are essential for diagnosing, localizing, and mitigating such errors.

\subsection*{Evaluation protocols}

With more AI agents being developed, frameworks for biologists and lay user evaluations need to assess axes of agent performance beyond accuracy. Evaluating AI agents requires an analysis of their theoretical capabilities and an assessment of practical implications, including ethical considerations, regulatory compliance, and the ability to integrate into discovery workflows. The challenge lies in developing evaluations that consider these diverse factors. Agents that integrate ML tools, particularly those developed by corporations, may undergo updates without prior notice to users. This poses challenges for reproducibility, as updates may alter the model's behavior or performance without researchers being aware. The scientific community needs transparent change logs and version control for agents, akin to practice in software development.

Existing evaluation frameworks consider either holistic evaluations \citep{HELM, mialon2023gaia} or benchmark the models for weak spots such as task framing \citep{srivastava2023imitation, Huang2023BenchmarkingLL}, long temporal dependencies, invalid formatting or refusal to follow instructions \citep{liu2023agentbench}. A caveat of such frameworks is the risk of evaluating how well the agents have learned to use specific APIs versus general results grounded in real-world interaction. Another challenge in evaluating agents is that biological systems are inherently dynamic, characterized by non-stationary distributions that evolve due to genetic mutations, environmental changes, and evolutionary pressures. Agents trained on static datasets may struggle to accurately model or predict outcomes in these changing systems. The challenge lies in developing agents capable of adapting to or continuously learning from new data, ensuring their predictions remain accurate as the underlying biological systems change. Techniques such as online learning, transfer learning, and reinforcement learning can be used to address this issue, but they come with their own set of challenges related to data availability and model complexity. Another challenge is the lack of standardization in biomedical discovery workflows, including data generation protocols that vary based on factors like disease cell lines, dosage levels, and time points~\cite{corsello2017drug}. This variability complicates the evaluation of agents for experimental planning. Evaluation of agents that use computational tools and databases will benefit from the increasing availability of standardized and application APIs~\cite{cohen2017scientific,lamprecht2020towards}.

\subsection*{Dataset generation}

As laid out, the vision for biomedical AI agents requires the capability of seeking, aggregating, perceiving, and reasoning over data from various modalities, created using differing specifications and with inherent variation in quality and volume. To support this vision, there is a critical need for large, open datasets that are both comprehensive and accessible, enabling the development of models across biological applications. Much human effort in building systems for biomedical research is dedicated to gathering and preparing such data for use in ML models (e.g., specific to a particular modality, such as graphs, time series, or discrete sequences \citep{ZITNIK201971}). This requires vetting processes and clear criteria for assessing the reliability and applicability of datasets.

Noisy data, characterized by errors, inconsistencies, and outliers, poses a significant challenge for models attempting to extract meaningful patterns and insights with minimal human oversight or data preparation effort. In addition, multi-modal data requires models to process different data representations and formats and bridge semantic gaps between them. Tackling these challenges necessitates advanced feature extraction, fusion, and noise mitigation techniques while maintaining robustness. As no pretraining phase (no matter how extensive) will be able to provide adequate examples from all data sources, models will also have to generalize to previously unseen sensory inputs.

\subsection*{Governance of AI agents}
\label{sec:ai_governance}
The governance of AI agents presents challenges that intersect technological, scientific, ethical, and regulatory domains. One challenge is establishing comprehensive governance frameworks that balance innovation with accountability \citep{WhiteHouse2023AIBillOfRights}. As AI agents gain autonomy, the necessity for robust guidelines to ensure responsible development, deployment, and commercialization grows. The discourse increasingly advocates for agent safeguarding to take precedence over further advancements in autonomy. Yet, navigating the regulatory landscape and forging an international consensus on AI governance remains complex while the advancement of agent capabilities continues. Striking a balance between innovation and safeguarding against potential risks requires collaboration among industry leaders, scientists, and policymakers \citep{Guha2023AIRegulation}. 

Safe adoption of AI agents requires addressing concerns of safe deployment. Aligning ML tools, such as LLMs, with ethical standards remains an open challenge, and ensuring the alignment of the agent as a digital entity raises complexity. Guidelines concerning human-agent interactions are underdeveloped despite the potential for unintended harmful consequences and malicious intent. 
Safeguarding frameworks are developed that include training, licensing, and mandatory safety and ethical compliance checks for agents~\citep{tang2024prioritizing}.

As AI agents become more integral to workflows in biological domains, monitoring their behavior grows increasingly complex. Currently, verifying the accuracy and trustworthiness of agent outputs is not straightforward, with only a limited number of systems capable of linking generated content to relevant references. 
It is essential to develop robust verification systems that can provide traceable references for generated content.
Assessing the synthesized knowledge may be impractical and unattainable as agents evolve further. 
When agents' capabilities become comparable to those of human experts, the risk of becoming overly reliant on AI increases, which could lead to a decrease in human expertise. In the worst-case scenario, such reliance could introduce a broad spectrum of safety hazards due to inadequate oversight.
To address these challenges, human-in-the-loop approaches can help maintain accountability. Continuous training and development of human expertise alongside AI can mitigate the risks of over-reliance on AI.

\subsection*{Risks and safeguards}

Autonomous experiments that do not include careful planning, broad consultation, competent execution, and ongoing adaptation might create long-term harms that outweigh the benefits. Although anticipating all potential complications is impossible, exploring possible problems early and frequently could reduce the expected cost of such issues.
The ethical and technical considerations relevant to AI agents are vast and deeply interconnected, particularly in biomedicine. This section will highlight some key categories.

Neglect can lead to risks similar to those of malicious intent. Multi-agent systems where some agents represent LLMs might, through equipment malfunctions and insufficient maintenance, inadvertently create harmful substances, for instance, by contaminating a procedure that would otherwise be safe. This issue is not unique to multi-agent systems; instead, it is a general lab safety concern. However, the absence of close human supervision removes a critical auditing layer. The increased role of automation in agent systems raises safety issues: a powerful, unaligned system prone to misinterpreting user requests or unfamiliar with lab safety practices could, given access to a well-stocked scientific facility, do damage by, for instance, mixing volatile substances or developing and dispersing toxins or pathogens. These are among the scenarios that most concern AI safety researchers.

Agents leverage LLMs' world knowledge and general reasoning abilities obtained during pretraining for robotics and planning. However, while efforts have been made to teach the robots the ``dos,'' the ``don'ts'' received less attention. Teaching robot agents the ``don'ts'' is crucial to convey instructions about prohibited actions, assessing the agent's understanding of these restrictions, and ensuring compliance~\cite{yang2023plug}. 
For LLM agents, plug-in safety chips~\cite{yang2023plug} feature safety constraint modules that translate natural language constraints into formal safety constraints for the robot to adhere to.
Experiments with robots highlight the potential for integrating formal methods with LLMs for robotic control.

LLMs trained in code completion can write Python programs from docstrings~\cite{chen2021evaluating} by training the model on the code completion task to write the code based on natural language commands~\cite{liang2023code}. Given natural language commands, these code-writing LLMs can be re-purposed to write robot policy code. However, if the translation inaccurately reflects the intended safety constraints, it could lead to either overly restrictive behavior, preventing the robot from performing its tasks effectively, or insufficiently stringent constraints, leading to safety violations. However, the robot policy code is less reliable for enforcing safety constraints than verifiable safe operations that satisfy standards such as ISO 61508. The approach assumes that all given instructions are feasible and lacks a mechanism to predict the correctness of a response before execution. However, due to their reliance on patterns in the training data, LLMs might generate syntactically correct but semantically inappropriate code. 
Additionally, generalizing plans across robotic embodiments is brittle with current LLMs.

Addressing the ethical implications of AI agents is paramount, given the direct impact on human and animal health and life. 
The handling of sensitive biological and medical data necessitates robust technological and regulatory measures to ensure security and confidentiality. One promising approach involves using privacy-preserving computation to train agents to protect the privacy of highly sensitive medical data. Homomorphic encryption can secure sensitive data by allowing computations on encrypted data and federated learning techniques allow training agents in a distributed manner without the need to centralize from across sites into a single data repository.

Algorithmic fairness is equally crucial, as biased AI agents can exacerbate health disparities across patients and increase inequalities in the volume of generated datasets and quality of biomedical knowledge, especially for diseases in long-tailed distributions in biological systems. The development of techniques such as adversarial debiasing and fair representation learning offers promising avenues to mitigate these risks.
In addition, the black-box nature of these compound AI systems poses another challenge, particularly in healthcare, where interpretability is vital for clinical adoption and patient trust. To provide clearer rationales for the agents’ decisions and make them more acceptable to users, it will become crucial to incorporate interactive dialogue systems that explain agentic outputs through natural language conversations.
Ethical considerations surrounding biosafety emerge as AI agents advance toward Level 3 agents. These issues intersect with ongoing debates in bioethics regarding synthetic biology, artificial organisms, and AI-driven life forms, requiring regulatory guidance and engagement from bioethicists and safety experts to ensure alignment with societal values and safety standards.

\subsection*{Challenges uniquely relevant for biomedical AI agents}
Biomedical AI agents face several unique challenges that distinguish them from other applications of AI. While strong AI agents have the potential to mitigate some of these challenges, their implementation in biomedical research requires careful consideration. One of the primary challenges is the need for robust and reliable systems capable of reasoning, planning, and executing actions in both virtual and hybrid virtual-physical environments. For instance, natural language reasoning chains can enhance the interpretability of an agent's actions and contextual outcomes, aiding researchers in understanding AI-generated insights. However, certain challenges persist that can delay the reliable implementation of AI agents or even cause harm if these systems are deployed prematurely. A critical issue is the difficulty in distinguishing between correlation and causality. Current AI agents often struggle with generating strong hypotheses, reasoning, and conducting experimental validations, tasks that typically require advanced AI systems (Level 3 agents) or human intervention. Moreover, AI agents need improved interfaces to interact safely and effectively with high-throughput experimental platforms. These platforms themselves face limitations in producing unbiased, AI-ready datasets that accurately capture the intra- and inter-variation inherent in biological systems. Such limitations hinder the generalization capabilities of AI agents, which rely on comprehensive and high-quality data to function optimally. The absence of data from high-throughput techniques can lead to AI agents forming false hypotheses or causing harm. This risk is exacerbated when AI agents work with small, biased biological datasets, which may be affected by issues like batch effects.

%% file: 07outlook.tex
\section*{Outlook}

Biomedical research is undergoing a transformative era with advances in computational intelligence. Presently, AI's role is constrained to assistive tools in low-stake and narrow tasks where scientists can review the results. We outline agent-based AI to pave the way for systems capable of reflective learning and reasoning that consist of LLM-based systems and other ML tools, experimental platforms, humans, or even combinations of them. 
The continual nature of human-AI interaction and building trustworthy sandboxes~\citep{schwartz2023enhancing}, where AI agents can fail and learn from their mistakes, is one way to achieve this. This involves developing AI agents proficient in various tasks, such as planning discovery workflows with machine learning feedback loops for experiments and performing self-assessment to identify and seek out gaps in their knowledge, fostering natural and artificial intelligence.

\subsection*{Ensuring context-appropriate and user-specific agent behavior}
To ensure agents behave as intended, it is essential to focus on their robustness and reliability by implementing evaluation protocols that test agents in diverse scenarios to identify potential vulnerabilities. Moreover, grounding agents in ethical guidelines and documentation, such as lab protocols and safety guidelines, is vital to align their actions with human values and safety standards. By addressing these aspects, we can ensure that the behavior of biomedical agents is both reliable and ethically compliant.

Concretely, we believe that in the early stages of technological adaptation, it is desirable to limit an agent's capabilities to a subset of their full potential by restricting action spaces, thereby eliminating the chance of catastrophic risk (e.g., decisions resulting in loss of life). Similar precedents for technological adaptation are already in place for other autonomous systems with similar risk profiles, such as autonomous driving, where a staggered technological adaptation is motivated by ethical considerations.

\subsection*{Governance and responsible human-AI partnership}

Managing errors requires designing strategies to diagnose, localize, and mitigate them. To diagnose errors internally, agents should use their reasoning abilities to build self-evaluation schemes, allowing them to assess their current status and actions. Externally, training independent anomaly detection and distribution shift models with domain knowledge of specific biomedical use cases can provide additional supervision to diagnose errors. Iterative agent interactions can result in cascading errors. To mitigate this, the evaluation agent can apply reverse reasoning chains to trace back to the initial error. Enhancing the adaptive reasoning abilities of agents is crucial for dynamically adjusting to changing conditions and rectifying errors as they occur.

In addressing the challenge of governance, it is our view that the required broad consensus is best reached in multi-disciplinary, cross-partisan, non-profit, and public institutions with the objective of public good. In this regard, we welcome the recent establishment of several public AI-focused safety institutions to facilitate that discussion. Our concrete advice for biomedical AI agents would be to establish focus groups or select committees with the required expertise that can define necessary ethical and technical evaluation standards, based on which concrete regulation can be determined (e.g., a necessary degree of human oversight and accountability structures). We furthermore advocate for standards and policies to be developed in the broadest international institutions possible to reduce the chance of risks simply being outsourced to jurisdictions with no existing or unenforceable regulations.

By fostering responsible human-AI partnerships and robust governance frameworks, we can harness the transformative potential of AI agents in biomedical research. This collaborative approach can pave the way for groundbreaking advances, eventually enhancing human health and well-being.

%% file: 08figure.tex
\begin{figure}
	\centering
	\includegraphics[width=\linewidth]{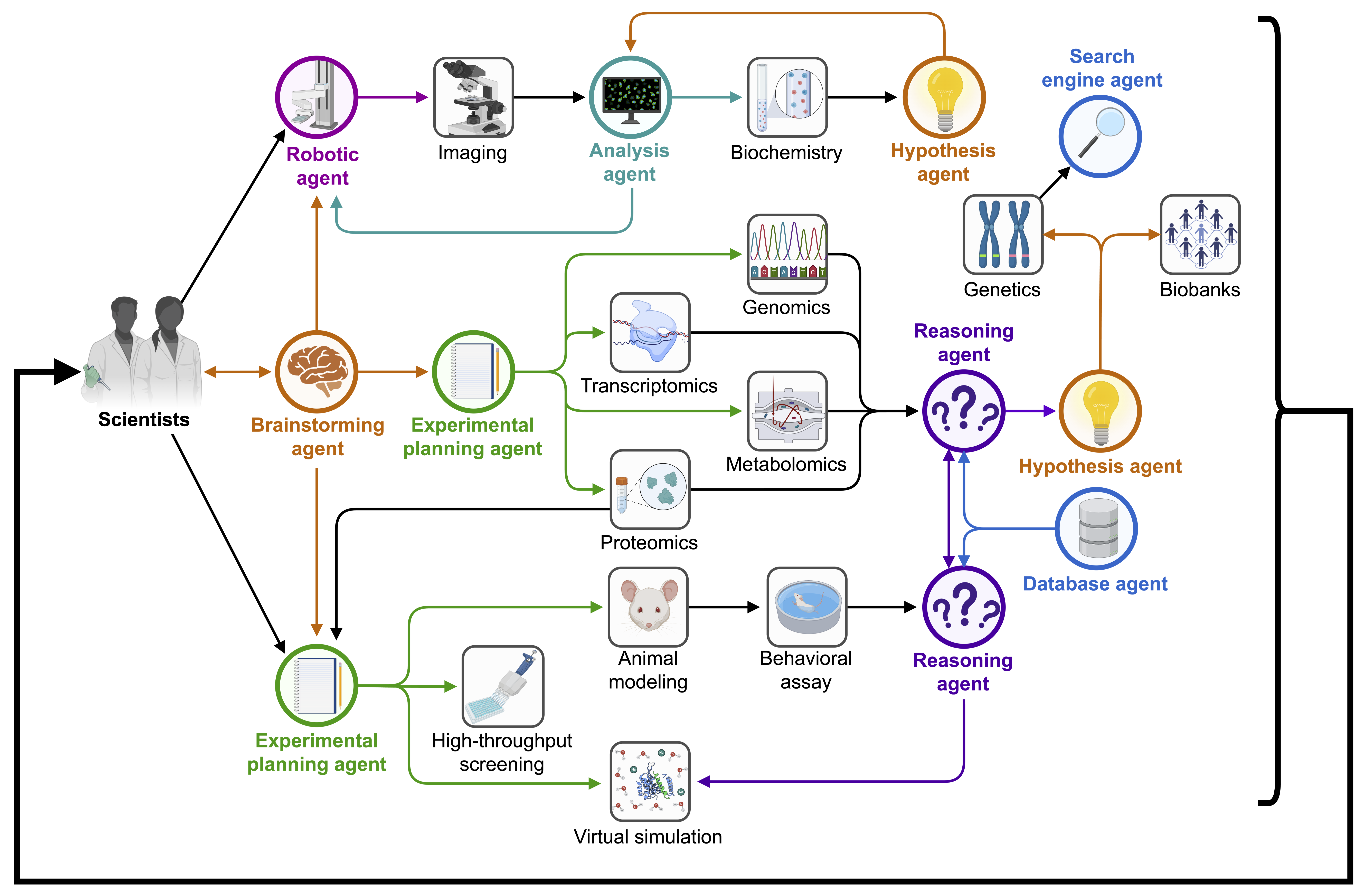}
	\caption{\textbf{Empowering biomedical research with AI agents.}
		AI agents pave the way for "AI scientists" capable of skeptical learning and reasoning. These multi-agent systems consist of agents based on conversable large language models (LLMs) and can coordinate machine learning (ML) tools, experimental platforms, humans, or even combinations of them. Robotic agent, AI agent that operates robotic hardware for physical experiments;
		Database agent, AI agent that can information in databases via `function calling' and APIs;
		Reasoning agent, AI agent capable of direct reasoning and reasoning with feedback;
		Hypothesis agent, AI agent that is creative and reflective when developing hypotheses, capable of characterizing its own uncertainty and using that as a driver to refine its scientific knowledge bases;
		Brainstorming agent, AI agent that generates a broad spectrum of research ideas;
		Search engine agent, AI agent that uses search engines as tools to rapidly gather information;
		Analysis agent, AI agent capable of analyzing experimental results to summarize findings and synthesize concepts;
		Experimental planning agent, AI agent that optimizes an experimental protocol for execution.
	}
	\label{fig:fig1}
\end{figure}

\clearpage

\begin{figure}
	\centering
	\includegraphics[width=\linewidth]{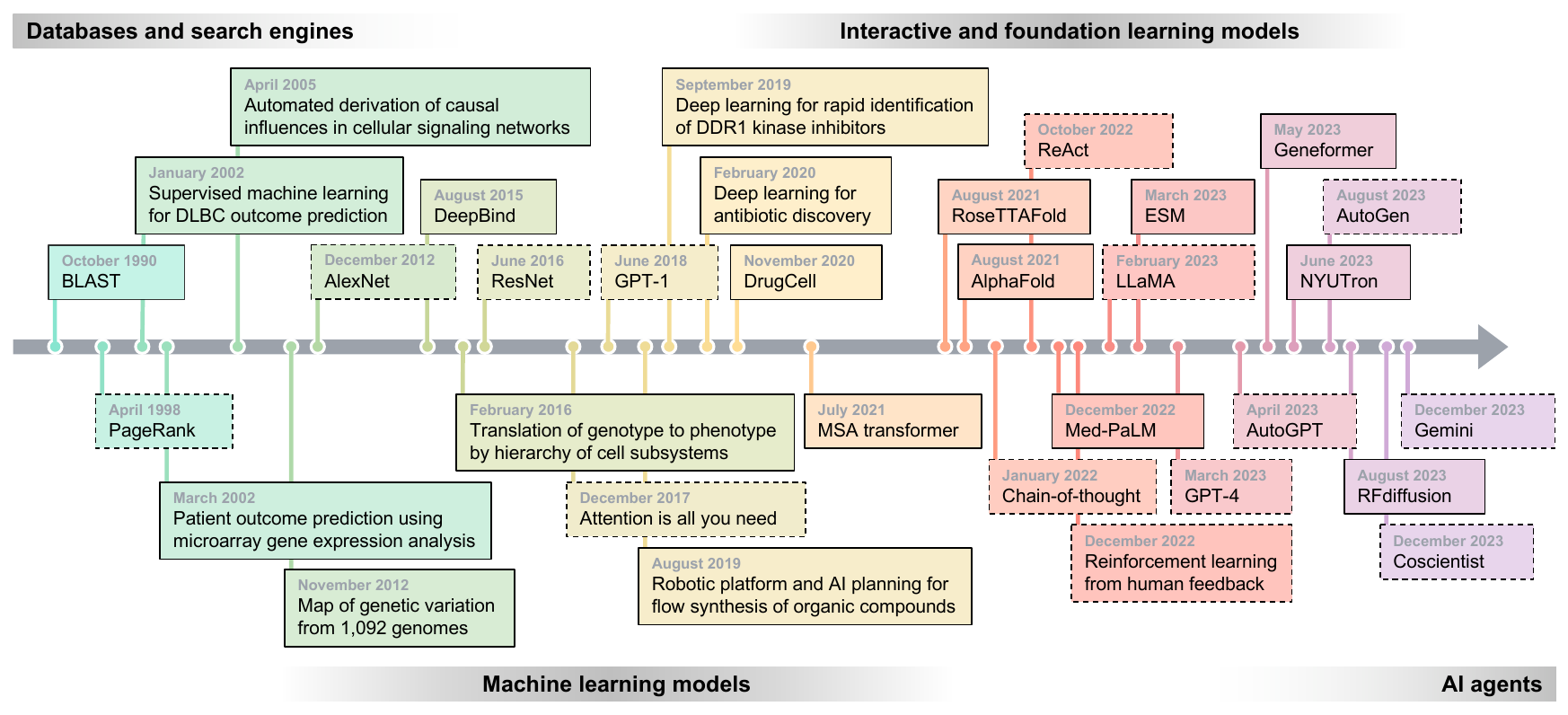}
	\caption{
		\textbf{Evolving use of data-driven models in research.}
		Data-driven approaches, from databases and search engines, machine learning, and interactive learning models to advanced agent systems (Section~\ref{sec:why}), have reshaped biomedical research throughout the last several decades. Dashed boxes represent studies focused predominantly on algorithmic machine learning innovation; solid-line boxes represent studies focused predominantly on biomedical discovery.
	}
	\label{fig:timeline}
\end{figure}

\clearpage

\begin{figure}
	\centering
	\includegraphics[width=\linewidth]{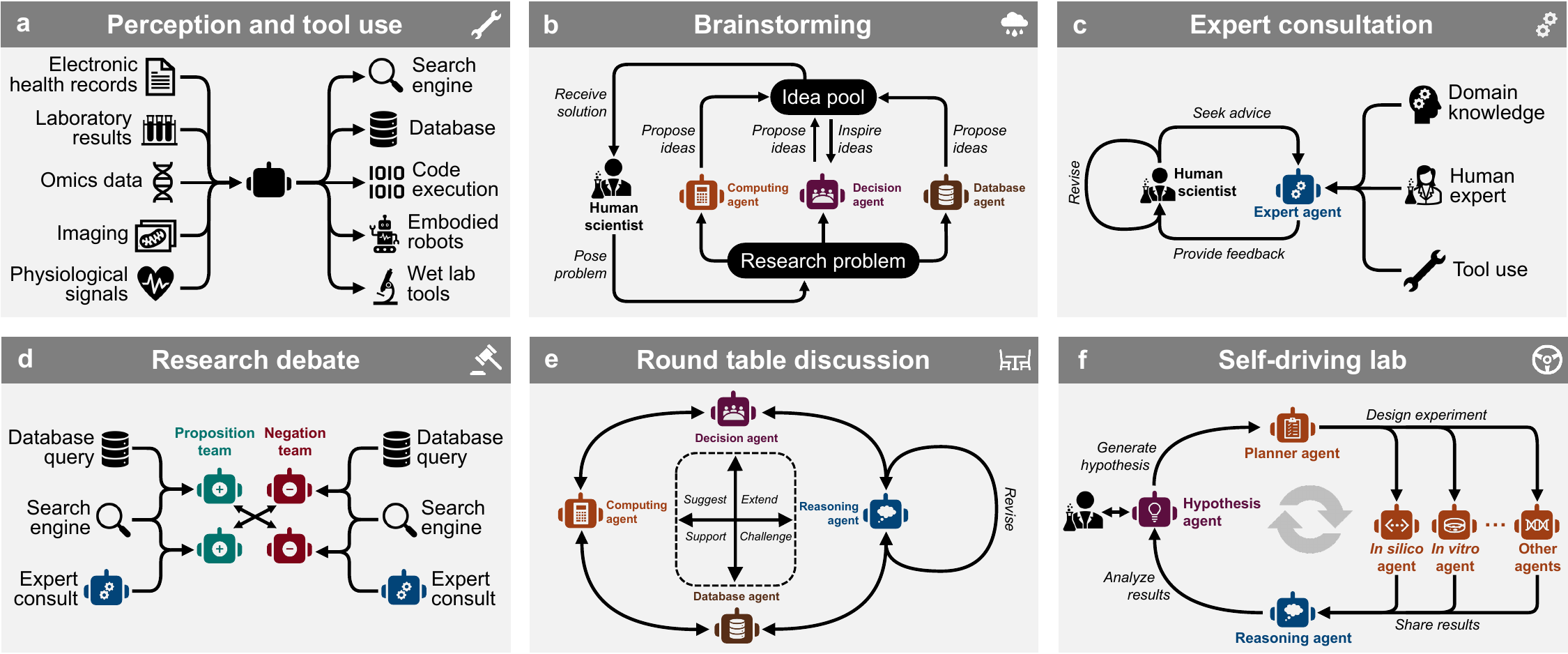}
	\caption{
		\textbf{Diverse configurations of AI agents in biomedicine --  from an LLM-based AI agent to a multi-agent system with AI models, tools, and integrated physical devices.}
		\textbf{a.} By programming an LLM with the role, one LLM-based agent, equipped with memory and reasoning abilities, performs multi-modal perception and utilizes a range of tools, e.g., web lab tools, to accomplish specified tasks.
		\textbf{b-e.} Leveraging AI agents equipped with diverse roles, perception modules, tools, and domain knowledge enables collaboration between agents and scientists. This collaboration can adopt various schemes, such as expert consultation, debate, brainstorming, and round table discussions. \textbf{f.}~Multi-agent systems can establish a self-driving laboratory wherein numerous agents collaborate on multiple iterations of biological research assisted by humans. Each cycle of research encompasses the generation of hypotheses, the design of experiments, the execution of experiments both in silico and in vitro, and the analysis of results. Computing agent, AI agent that utilizes computational models as tools;
		Decision agent, AI agent that makes decisions in response to given conditions;
		Database agent, AI agent that retrieves relevant information from databases;
		Reasoning agent, AI agent capable of direct reasoning and reasoning with feedback;
		Expert agent, AI agent that provides professional consultation based on reliable sources, such as domain expertise, feedback from human experts, and the results of specific tools.
		Hypothesis agent, AI agent capable of reflective learning and reasoning to generate hypotheses;
		Planner agent, AI agent that devises plans for future actions;
		In silico/vitro agent, AI agent that uses tools in silico or in vitro environment.
	}
	\label{fig:achive_agent}
\end{figure}

\clearpage

\begin{figure}
	\centering
	\includegraphics[width=\linewidth]{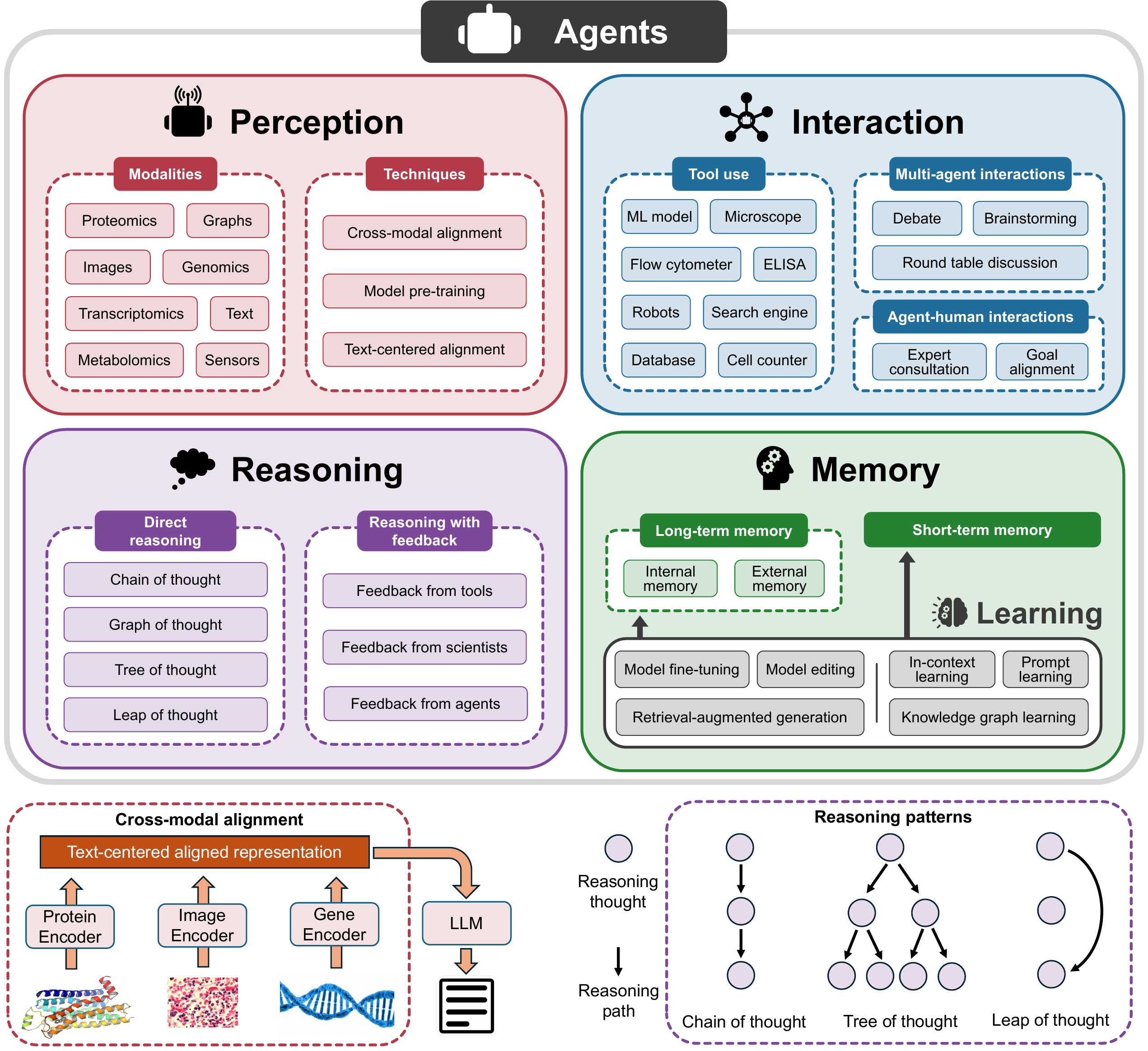}
	\caption{\textbf{Four key modules of biomedical AI agents: perception, interaction, reasoning, and memory modules.}
		Perception interprets multi-modal environmental data.
		Interaction facilitates engagement with the environment, encompassing human-agent interactions, multi-agent interactions, and tool use.
		Memory is responsible for the storage and retrieval of knowledge,
		while Learning focuses on the acquisition and updating of knowledge.
		Reasoning, with or without environmental feedback, plays a crucial role in planning and decision-making processes.
		Cross-modal alignment is a key technique for the perception of LLM-based agents, where inputs from different modalities are aligned within a text-centered representation space. This alignment enables the LLM to perceive and process various input modalities.
		Reasoning patterns for AI agents indicate transitions between reasoning thoughts. For instance, agents with a chain of thought pattern generate reasoning in a step-by-step manner.
	}
	\label{fig:componenet_agent}
\end{figure}

\clearpage

\clearpage

\begin{figure}
	\centering
	\includegraphics[width=\linewidth]{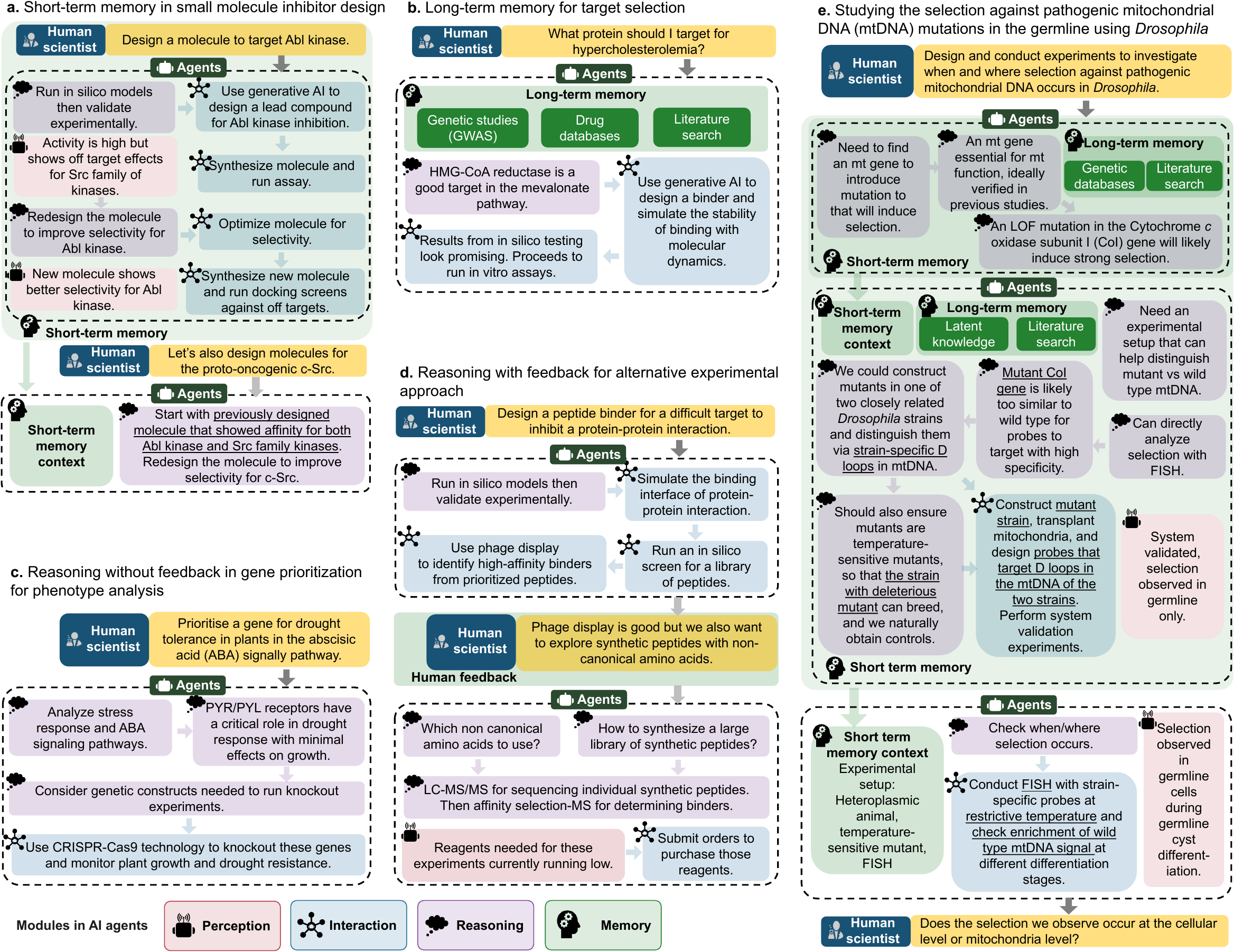}
	\caption{\textbf{Illustration of components in biomedical AI agents.}
		\textbf{a.} Use of a short-term memory module to recall previous relevant experiments for small molecule inhibitor design.
		\textbf{b.} Use of a long-term memory module to retrieve relevant information for target selection for a disease.
		\textbf{c.} Use of reasoning without scientist feedback in gene prioritization for phenotype analysis.
		\textbf{d.} Use of reasoning with feedback from scientists to select an alternative experimental approach.
		\textbf{e.} Combining perception, interaction, memory, and reasoning modules to study the selection against pathogenic mitochondria DNA in the germline. }
	\label{fig:bio_agent_example}
\end{figure}

\clearpage

\begin{figure}
	\centering
	\includegraphics[width=\textwidth]{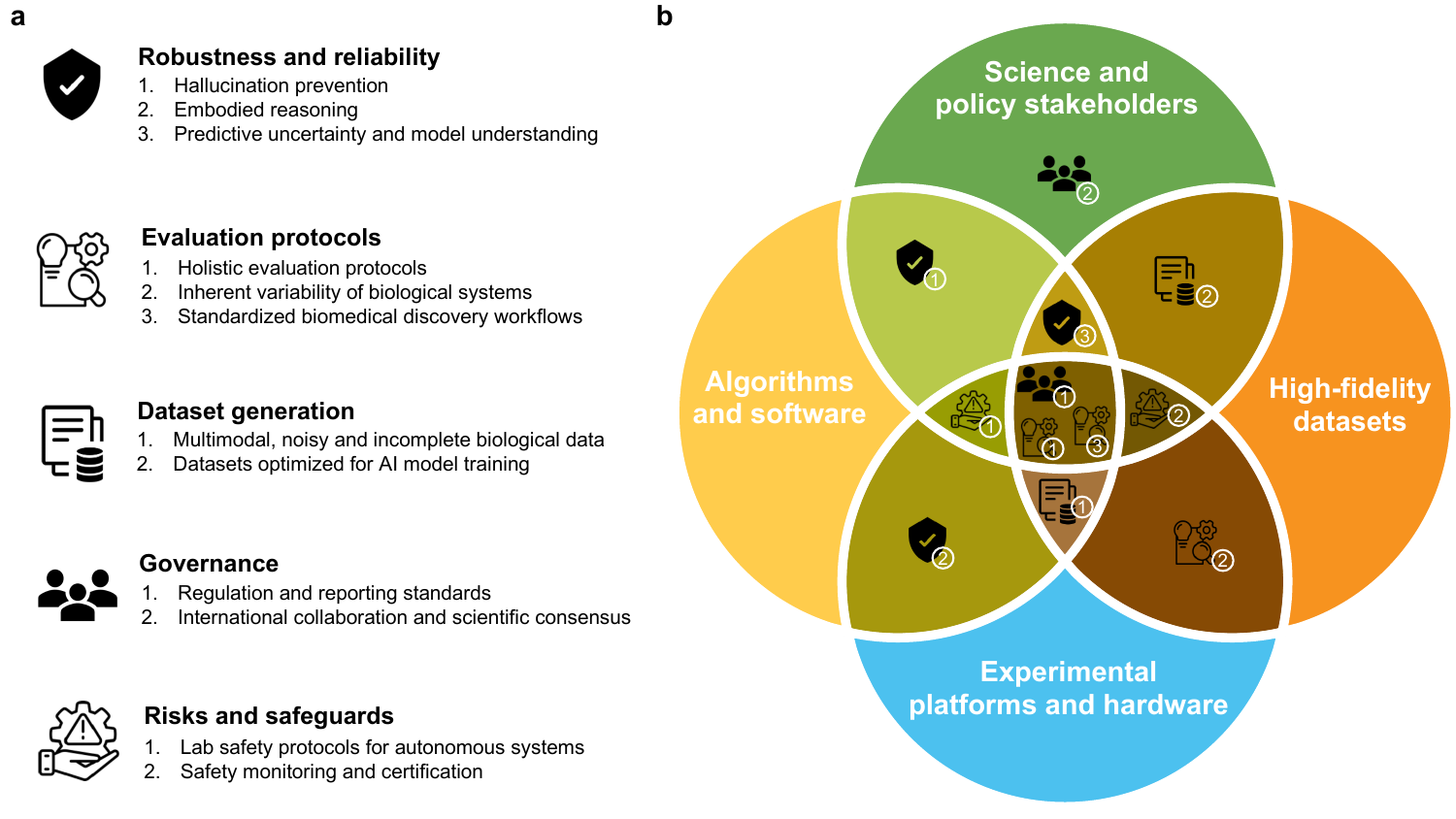}
	\caption{\textbf{Challenges for AI agents in biomedical discovery.} \textbf{a.} Shown are critical challenges--including robustness and reliability, evaluation protocols, dataset generation, governance, and risks---alongside \textbf{b.} strategic approaches to address them. }
	\label{fig:challenges}
\end{figure}

%% file: 09box.tex
\begin{table}[h!]
    \centering
    \footnotesize
    \setlength{\tabcolsep}{5pt}
    \begin{tabular}{>{\arraybackslash}p{2.1cm}||>{\raggedright\arraybackslash}p{3cm}|>{\raggedright\arraybackslash}p{3.3cm}|>{\raggedright\arraybackslash}p{3.3cm}|>
    {\raggedright\arraybackslash}p{4cm}}
        \textbf{Autonomy} & \multicolumn{3}{c|}{\textbf{Biomedical discovery}}  & \multirow{ 2}{*}{\textbf{Scientist-AI agent roles}} \\
        \textbf{levels} &  \multicolumn{1}{c}{\textbf{Hypothesis}} & \multicolumn{1}{c}{\textbf{Experiment}} & \multicolumn{1}{c|}{\textbf{Reasoning}} & \\
        \toprule \rowcolor{SkyBlue!5}
        \textbf{Level 0:} \par\vspace{3pt} \textbf{No AI agent} & None & ML models perform predefined tasks, with no adaptive changes to the protocols & None & $\cdot$ Scientist defines the hypothesis and sometimes uses the output of ML models to help with their generation \par\vspace{3pt}
        $\cdot$ Scientist defines the task to test hypothesis \par\vspace{3pt}
        $\cdot$ Scientist completes tasks\\
        \midrule \rowcolor{SkyBlue!15}
        \textbf{Level 1:} \par\vspace{3pt} \textbf{AI agent as an} \par \textbf{assistant} & AI agent formulates simple and narrow hypotheses that are a direct composition of existing knowledge, preliminary data, or observations & Narrow design of experimental protocols and utilization of in silico and experimental tools & $\cdot$ Strong reasoning in a selected task \par\vspace{3pt} $\cdot$ Multi-modal summary of findings \par\vspace{3pt} $\cdot$ Use of experimental data and existing knowledge & $\cdot$ Scientist defines the hypothesis \par\vspace{3pt} $\cdot$ Scientist defines the series of tasks to test hypothesis \par\vspace{3pt}
       $\cdot$  AI agent completes tasks \par \\
        \midrule \rowcolor{SkyBlue!25}
        \textbf{Level 2:} \par\vspace{3pt} \textbf{AI agent as a} \par \textbf{collaborator} & AI agent generates hypotheses that are an explicit continuation of data trends and known literature & $\cdot$ Design of rigorous experimental protocols and adept utilization of a broad range of ex silico tools \par\vspace{3pt} $\cdot$ Once data is collected, employ statistical and computational methods to analyze the results and interpret the data to determine whether it supports or refutes the hypothesis & $\cdot$ Interpreting findings within existing knowledge, considering alternative explanations, and assessing the reliability and validity of the findings \par\vspace{3pt} $\cdot$ Synthesis of concepts beyond a summary of findings  \par\vspace{3pt} $\cdot$ Collaborating with other researchers and undergoing peer review to validate findings and ensures that conclusions are robust and credible  & $\cdot$ Scientist proposes initial hypothesis and refines hypothesis with AI agent \par\vspace{3pt} $\cdot$ AI agent defines the series of tasks to test hypothesis \par\vspace{3pt} $\cdot$ AI agent completes tasks \par \\
        \midrule \rowcolor{SkyBlue!35}
        \textbf{Level 3:} \par\vspace{3pt} \textbf{AI agent as a} \par \textbf{scientist} & AI agent generates creative, de novo hypotheses that are indirect extrapolations from existing knowledge. & $\cdot$ Development of experimental methods unlocking new capabilities \par\vspace{3pt} $\cdot$ Actively gather data through experiments, observations, or simulations using various techniques and tools to measure and record biological phenomena accurately & $\cdot$ Based on the results and interpretations, refine and experimental approaches for continuous learning and adaptation to improve the accuracy and depth of understanding \par\vspace{3pt} $\cdot$ Concise, informative and clear conceptual links between findings  & $\cdot$ Scientist and AI agent together form hypothesis  \par\vspace{3pt} $\cdot$ AI agent defines the series of tasks to test hypothesis \par\vspace{3pt} $\cdot$ AI agent completes tasks \\
        \bottomrule
    \end{tabular}
    \caption{\textbf{Levels of autonomy in AI agents.} AI agents are characterized by four levels of autonomy in biological research, which are defined based on the capabilities of AI agents to complete different steps of the discovery process. At Level 0, there is no AI agent, and ML is used as a tool. Level 1 consists of AI agents as research assistants, where agents complete a set of narrow and specific tasks defined by scientists. At Level 2, AI agents act as collaborators and can use a broad set of tools to identify scientific discoveries. Still, they can only generate hypotheses that are a linear continuation of literature. Finally, at Level 3, AI agents act similarly to human scientists across several axes of human evaluation, capable of identifying and understanding pioneering discoveries and extrapolating novel hypotheses that cannot be derived from existing knowledge.
    }
    \label{tab:levels_table}
\end{table}

\clearpage
\begin{table}[h!]
    \centering
    \small
    \setlength{\tabcolsep}{5pt}
    \begin{tabular}{>{\arraybackslash}p{2cm}||>{\raggedright\arraybackslash}p{4cm}|>{\raggedright\arraybackslash}p{6cm}|>{\raggedright\arraybackslash}p{4cm}}
         \textbf{Autonomy levels} & \textbf{Genetics (mutational effect modeling)} & \textbf{Cell biology (drug resistance)}  & \textbf{Chemical biology (binder design)}\\
        \toprule \rowcolor{SkyBlue!5}
        \textbf{Level 0} & Statistical package to analyze a pre-selected GWAS study. & Use of ML tools for modeling cellular outcomes of drug perturbations, including cell imaging, omics, and viability. & Use of ML tools for protein structure prediction, molecular docking, and generative models for binder design. \\
        \midrule \rowcolor{SkyBlue!15}
        \textbf{Level 1} & To explore potential mutational associations with disease, writes bioinformatics software for quality control and statistical analysis of genotype data from pre-fetched relevant GWAS studies. & Integrates multimodal (imaging, omics, viability) and multiscale (cellular, tissue) data to create in silico models of drug resistance. Retrieves and executes existing experimental protocols to study resistance. Analyzes raw image and omics data with predefined pipelines. & Studies a specific protein target, integrates ML tools, such as AlphaFold for structure prediction and neural networks for screening chemical libraries to find candidate chemical compounds to bind to the target. \\
        \midrule \rowcolor{SkyBlue!25}
        \textbf{Level 2} & Selects GWAS studies relevant to a provided hypothesis. If none exists, it designs and executes its own study or pulls other relevant genomic data to investigate the hypothesis. & Autonomously develops and adaptively refines hypotheses about resistance mechanisms based on knowledge and real-time experimental data analytics. Designs and executes scalable and cost-effective experimental protocols with experts in the loop. & Designs binders for more challenging targets. Identifies scaffolds that bind to similar pockets and adapts them for the target. Synthesizes and tests molecules using existing experimental techniques.\\
        \midrule \rowcolor{SkyBlue!35}
        \textbf{Level 3} & Initiates genomic studies and optimizes non-invasive methods of DNA collection for cost-effectiveness and ease of participant requirement. Innovates statistical methods to identify causal variants from genotypic data and develops in vitro techniques for validating candidate gene markers in disease models. & Proactively identifies critical unresolved problems in drug resistance, proposing innovative therapeutic strategies. Performs in silico simulations of cellular dynamics in tumor contexts and under complex perturbations (combinatorial genetic and chemical perturbations under different dosing schedules). Develops novel highly multiplexed in vivo single-cell spatial technologies, enabling live tracking of gene expression, molecular interactions, and cell-cell interactions during resistance evolution. & Proposes de novo binders for an undruggable target or a poorly studied target. Designs in situ experiments to study molecular interactions. Synthesizes molecules with more complex pathways and designs and executes assays to test efficacy.
 \\
        \hline
    \end{tabular}
    \caption{\textbf{Examples of levels of autonomy of AI agents in genetics, cell biology, and chemical biology.}}
    \label{tab:sample}
\end{table}

\clearpage

\begin{longtable}{p{0.24\columnwidth}|p{0.75\columnwidth}}
\textbf{Term} & \textbf{Description}  
\\\midrule 
Multi-modal foundation model &
Advanced algorithms trained on multimodal datasets that can process various data types, including text, images, biological sequences, and high-dimensional tabular readouts. This training allows them to perform a broad array of tasks through few-shot fine-tuning and prompting across domains with little to no additional training \\ \midrule
Transformer architecture  & Deep learning model architecture that uses on self-attention mechanism to capture long-range dependencies in input sequence data
\\ \midrule
Large language model  & 
Machine learning model with parameters on the scale of billions, trained on vast amounts of text data to understand, generate, and interact with human language on a large scale\\ \midrule
Generative pretraining & 
Strategy for training a machine learning model in an autoregressive manner to predict the next token from given data tokens, facilitating a general understanding of data sequence likelihoods 
\\ \midrule
LLM-based AI agent & AI system capable of solving complex tasks within its environment by equipping large language model with modules for perception, interaction, memory, and reasoning   \\  \midrule
Embodied AI agent & AI agent system that interacts with the physical world through a body. The embodiment enables the agent to learn and adapt from sensory feedback and physical interactions \\  \midrule
Fine-tuning & 
A training process of making small adjustments to a pre-trained machine learning model to improve its accuracy on a specific task or dataset \\  \midrule
Instruction tuning & 
A training strategy that fine-tunes a model using a dataset of instructions and corresponding outputs to enhance its ability to follow specific instructions\\  \midrule
Reinforcement learning with human
feedback & 
A reinforcement learning strategy where an action model learns to perform tasks by receiving feedback from a reward model that mimics human preferences, guiding it to align with desired human behaviors  \\ \midrule
Prompting & Techniques that provide specific text or other modal input instructions to guide the model in responding toward a desired output direction\\ \midrule
Cross-modal alignment & A training scheme to align the representation embeddings of models across various modalities\\ \midrule
In-context learning & 
Ability to perform new tasks based on a handful of examples provided within the contextual prompt, without requiring explicit model training \\ \midrule
Retrieval-augmented generation & Techniques that make generative models to produce contextually relevant text by retrieving pertinent information and using it to inform the generation process
\\\bottomrule
\caption{\textbf{Glossary of key machine learning terms.}
\label{table:glossary_ml}}\\
\end{longtable}

\clearpage

\begin{longtable}{p{0.24\columnwidth}|p{0.75\columnwidth}}
\textbf{Term} & \textbf{Description}  
\\\midrule 
Linkage disequilibrium  & A phenomenon in which two alleles occur so often in proximity in the chromosome that their association cannot be random \\
\midrule 
Single-nucleotide \newline polymorphisms & Genetic variation consisting of the replacement of a single nucleotide in the DNA sequence     \\
\midrule 
Genome-wide association study &  Approach that identifies genetic variations across the entire genome associated with a specific disease or complex trait \\
\midrule 
Pharmacogenetics  &  Field of research that aims to understand individuals' responses to different drugs based on their genetic factors \\
\midrule
Experiment in-vitro  &  Procedures and investigations that occur within a laboratory environment (e.g., in a test tube) and outside of living organisms \\
\midrule
In silico modeling & The use of computers to build simulations or experiments that recreate complex biological phenomena in order to be able to study and predict specific behaviors. For example, modeling of molecular dynamics\\
\midrule
Mass spectrometry  &  Analytical tools to characterize and identify individual molecules based on specific properties (e.g., mass-to-charge ratio) \\
\midrule
Molecular docking & Computational simulation tools used to predict how ligands bind to receptors   \\
\midrule
Retro-synthesis &  Techniques to design the synthesis of complex molecules by starting from the target and moving back to the original compounds  \\
\midrule
Crystallography &  Field of science studying the structure of atoms and molecules in crystals, which are solid materials whose compounds are ordered according to a very regular and ordered arrangement \\
\midrule
Cryo-electron \newline microscopy  & Imaging techniques used to identify the 3D structure of bio-molecules with near-atomic resolution without the need for extensive sample preparation and with the overall preservation of the sample 
\\\bottomrule
\caption{\textbf{Glossary of key biological terms.}
\label{table:glossary_bio}}\\
\end{longtable}